\documentclass[lettersize,journal]{IEEEtran}
\usepackage{amsmath,amsfonts}
\usepackage{algorithmic}
\usepackage{algorithm}
\usepackage{array}
\usepackage[caption=false,font=normalsize,labelfont=sf,textfont=sf]{subfig}
\usepackage{textcomp}
\usepackage{stfloats}
\usepackage{url}
\usepackage{verbatim}
\usepackage{graphicx}
\usepackage{cite}

\usepackage{multirow}
\usepackage{amssymb}
\usepackage{enumitem}
\usepackage{makecell}

\begin{document}

% \title{Locker-Based Truck–Drone Routing with Pickup and Delivery Considering No-Fly Zones}
\title{Locker-based Truck-Drone Routing with Integrated Considerations of Pickups, Deliveries, and No-Fly Zones}

\author{Xuanyu Liu, Hui Hu*, Jiao Zhao, Ziliang Wang*, Zhengbing He, {\it Senier Member, IEEE}
        % <-this % stops a space
% 脚注
\thanks{This work was supported by the National Natural Science Foundation of China (72274024), the Shaanxi Provincial Key R\&D Program, China (2025GH-YBXM-027, 2024CY2-GJHX-31). *Corresponding: H. Hu (huhui@chd.edu.cn) and Z. Wang (wangziliang@chd.edu.cn).}
\thanks{X. Liu, H. Hu, J. Zhao, and Z. Wang are with the College of Transportation Engineering, Chang'an University, China; 
Z. He is with the Faculty of Science and Technology, University of Nottingham Ningbo China. }
}

\markboth{ }%
{Shell \MakeLowercase{\textit{et al.}}: A Sample Article Using IEEEtran.cls for IEEE Journals}


\maketitle
\begin{abstract}
Truck-drone delivery is an emerging last-mile logistics mode combining the long-haul capacity of trucks with the flexible service capability of drones. In locker-based operations, smart lockers serve not only as temporary parcel storage facilities but also as automated drone docking and service nodes. These automated nodes support drone takeoff, landing, parcel handover, and battery replacement, thereby significantly extending the service range and operational flexibility of drone-assisted delivery networks. However, practical locker-based delivery systems face complex real-world challenges, requiring the integrated coordination of not only parcel delivery, return pickup, battery-constrained and load-dependent drone flights, but also necessary detours around restricted airspace. To address this practical and multifaceted challenge, this paper introduces a locker-based truck-drone routing problem with integrated considerations of pickups, deliveries, and no-fly zones (LTDRP-PDNF), with the objective of minimizing the total operational cost of a fleet of drone-equipped trucks. We formulate the route construction process as a Markov Decision Process and develop a two-stage deep reinforcement learning-based neural heuristic. The first stage utilizes an attention-based encoder and a Bidirectional Gated Recurrent Unit decoder to solve the truck-only routing problem, formulated as a capacitated vehicle routing problem. The second stage combines a policy-transfer strategy with a hybrid dispatch assignment heuristic to construct fully coordinated truck and drone routes for LTDRP-PDNF.  Experiments on instances of different scales demonstrate that the proposed method outperforms metaheuristic and neural heuristic baselines in most cases while maintaining exceptionally short computation times, offering an effective, scalable solution framework under practical operational constraints.

\end{abstract}

\begin{IEEEkeywords}
Low-Attitude Economy, Drone, Deep Reinforcement Learning, Logistics, Vehicle Routing Problem.
\end{IEEEkeywords}

\section{Introduction}

\IEEEPARstart{D}{riven} by the explosive growth of online retail and increasingly exacting customer demands, last-mile delivery systems are facing growing pressure to provide faster, more flexible, and more cost-efficient services in complex urban transportation environments and remote delivery areas \cite{Duan2025}. The use of unmanned aerial vehicles (UAVs), or drones, for delivery represents an emerging direction in last-mile logistics, offering advantages such as rapid transit times, economical operation, and independence from road conditions. The commercial landscape for this novel offering is forecast to achieve a value of \$100 billion over the coming decade \cite{Liu2022}.

Although {\bf drone delivery} offers numerous advantages, it also faces several practical limitations, including restricted payload capacity, limited battery endurance, and susceptibility to volatile weather conditions \cite{Kong2024}. Additionally, {\bf no-fly zones}, such as airports, military bases, and prisons, impose strict spatial restrictions on drone operations, necessitating careful route planning to ensure safe and feasible delivery activities.

To combine the strengths of trucks and drones, {\bf truck-drone collaborative delivery} has been widely recognized as an effective logistics mode, which can improve service accessibility, reduce delivery time, and enhance the overall efficiency of logistics resource (see Section \ref{sec:Literature_Truck_Drone} for the literature). 

In such systems, trucks are better suited for long-haul transportation because of their higher payload capacity and extended operating range, whereas drones are particularly effective for short-distance last-mile services due to their flexibility and speed.

Meanwhile, {\bf locker-based delivery} has emerged as an innovative logistics model and has been increasingly adopted by logistics and e-commerce companies (see Section \ref{sec:Literature_Locker} for the literature). 
Several enterprises have explored this drone-locker integration mode. For example, Fengyi Logistics utilizes drone docking cabinets for package distribution, while Meituan employs drones in conjunction with smart lockers for food delivery services \cite{Hong2025}. In this delivery mode, lockers can serve not only as temporary storage facilities for parcels but also as drone service nodes that support takeoff, landing, parcel handover, and battery replacement. In practical operations, lockers may involve multi-product delivery demands and simultaneous pickup requirements, such as package returns, recycling, and reverse logistics services. Drones operating between trucks and lockers are subject to load-dependent energy consumption and detour requirements caused by no-fly zones.

From an optimization perspective, these operational features fundamentally change the decision structure of the routing problem. The planner must jointly optimize the truck visiting sequence, drone service assignment, truck–drone synchronization, and trip feasibility under payload, battery, and no-fly-zone constraints.
Neglecting any of these factors may render the resulting solution infeasible or ineffective in real-world operations.
Unfortunately, directly applying conventional vehicle routing models or standard truck–drone routing models may fail to capture the coupled decisions required in locker-based truck–drone delivery systems.

Motivated by the above practical need, this paper introduces a {\bf locker-based truck-drone routing problem with integrated consideration of pickups, deliveries, and no-fly zones}, referred to as LTDRP-PDNF. 
{\it 
In this problem, a fleet of drone-equipped trucks departs from a central depot to serve delivery stations and lockers. Trucks provide direct services to some locations and also launch or recover drones, while drones are assigned to serve lockers with multi-product pickup and delivery demands. During drone operations, battery replacement at trucks and lockers, load-dependent energy consumption, and no-fly-zone detours are explicitly considered to ensure operational feasibility. }

LTDRP-PDNF is computationally challenging because it requires the simultaneous construction of truck routes, drone service routes, locker service sequences, and energy-feasible flight paths. 
To overcome the challenge, we formulate the sequential route construction process of LTDRP-PDNF as a Markov Decision Process (MDP) and develop a two-stage deep reinforcement learning (DRL)-based neural heuristic to solve the problem. In the first stage, an attention-based encoder and a Bi-GRU decoder are employed to solve the truck-only routing problem, which is represented as a capacitated vehicle routing problem (CVRP). In the second stage, a transfer strategy combined with a hybrid dispatch assignment heuristic is designed to adapt the learned truck-routing policy to the more complex LTDRP-PDNF and construct coordinated truck-drone solutions.
This proposed solution outperforms both exact optimization methods and traditional heuristics and metaheuristics.
Specifically, exact optimization methods may provide high-quality solutions for small-scale routing problems, but their computational burden increases rapidly as the number of customers, lockers, and operational constraints grows. 
Traditional heuristics and metaheuristics can improve computational efficiency, but they often rely on handcrafted rules and problem-specific operators, which may limit their adaptability to structurally complex routing scenarios.

Comprehensive numerical experiments are conducted to evaluate the proposed method on both the CVRP and LTDRP-PDNF instances with different problem scales. The results demonstrate that the proposed method can generate high-quality solutions within short computation times and outperforms conventional heuristic, metaheuristic, and neural heuristic baselines in most tested cases. The results also show that the proposed method exhibits desirable scalability and generalization capability across varying problem sizes.

The remainder of the paper is organized as follows. Section II provides an overview of the relevant literature. Section III describes LTDRP-PDNF in detail. Section IV presents the MDP formulation. Section V introduces the proposed two-stage solution method. Section VI reports and analyzes the numerical experimental results. Section VII concludes the paper and discusses future research directions.

\section{Related Works}

\subsection{Collaborative truck–drone delivery problems}\label{sec:Literature_Truck_Drone}

In recent years, drones are increasingly being explored as an effective solution for delivery tasks due to their rapid flight capabilities and low operational costs, attracting significant research interest. The truck–drone routing problem was initially proposed in \cite{Murray2015}, where the authors investigated the flying sidekick traveling salesman problem (FSTSP). In the FSTSP, a variant of the classic traveling salesman problem, a truck equipped with a drone is utilized to serve all customers within a delivery network. \cite{Jeong2019} addressed the FSTSP with energy constraints and no-fly zones (FSTSP-ECNZ), developing a two-phase heuristic approach to enhance computational efficiency. 

Recent studies have explored increasingly sophisticated and efficient delivery models. For example, \cite{Xiao2024} developed a green vehicle routing problem with drones that considers the presence of steep roads (GVRPD-SR). The study introduced an improved adaptive large neighborhood search (IALNS) algorithm to solve the proposed model, in which the dynamic energy consumption of trucks and drones is taken into account. \cite{Mulumba2024} addressed the drone-assisted pickup and delivery problem, and employed an adaptive large neighborhood search metaheuristic to solve practical-sized instances. \cite{Meng2023} studied a multi-visit drone routing problem with pickup and delivery (MDRP-PD), considering the effects of both payload and travel distance on drone energy consumption, in addition to addressing customer pickup demands. \cite{Liu2023} provided a two-stage heuristic algorithm based on the simulated annealing algorithm to address the multi-visit drone-vehicle routing problem with simultaneous pickup and delivery considering no-fly zones (MDVRPSPDNF). \cite{Duan2025a} investigated the time-dependent multi-objective collaborative routing problem between a vehicle and a drone, one that incorporates concurrent pickup and delivery operations, and the authors developed an improved local search algorithm to effectively address this problem.

% B. Locker-based delivery
\subsection{Locker-based delivery}\label{sec:Literature_Locker}

In previous research on the use of lockers in delivery activities, their primary applications fall into two categories. The first category views lockers as intermediate storage points for trucks or drones, where parcels can be deposited and later retrieved by customers at their convenience. For instance, \cite{DellAmico2023} studied the pickup and delivery vehicle routing problem with lockers and time windows (PDVRPLTW), in which customers are given the option to choose among home delivery by truck, self-service collection from a nearby locker, or delegating the delivery method to the discretion of the logistics provider. Similarly,  \cite{Boschetti2024} solved the TSP with drone and lockers (TSPDL), which involves combining drones and trucks for efficient package delivery to either customers' homes or lockers. The second category treats lockers as mobile delivery units capable of autonomously performing last-mile delivery tasks within an integrated transportation system. For example, \cite{Li2021} integrated autonomous mobile lockers (AMLs) into existing urban logistics networks and introduced the AML-Courier 2-echelon location routing problem (AC-2E-LRP). In the proposed system, couriers collaborate with AMLs, which travel to meet them in the field to facilitate the exchange of parcels between couriers and the central depot.  \cite{Ensafian2023} examined the coordination between AMLs and couriers in the context of simultaneous pickup and delivery operations. Their model permits both AMLs and couriers to operate on open routes when such flexibility  contributes to overall cost reduction.

Recent technological advancements have driven the increasing adoption of intelligent delivery lockers, featuring highly automated systems, in logistics and distribution operations. These smart lockers can autonomously manage parcel handling, facilitating automated collection and dispatch. Additionally, their rooftops are designed to support drone takeoffs and landings, enabling effective integration into multi-modal distribution networks. Furthermore, integrated battery charging stations extend the operational range of delivery drones through mid-route energy replenishment, significantly boosting overall logistics efficiency. \cite{Zhu2024a} formulated the locker-based drone delivery makespan minimization problem (LDDMMP), where drones perform delivery tasks by flying between lockers, each equipped to facilitate battery replacement. To address this problem, the study developed a variable neighborhood search algorithm (VNS) and validated its effectiveness on practical instances. 
\cite{Zhu2024b} investigated the problem of minimizing the total travel distance in locker-based drone delivery, utilizing locker rooftops as drone parking platforms. The study developed a two-stage method, employing a branch-and-cut algorithm and a heuristic, to solve the hybrid drone routing and parking problem.

\begin{table*}[!t]
\caption{Comparison Between Related Studies And Our Study}
\label{table1}
\centering
\begin{tabular*}{\linewidth}{@{\extracolsep{\fill}} llcccccl @{}}
\hline
\multirow{2}{*}{References} & \multirow{2}{*}{Problem} & \multicolumn{5}{c}{Problem Characteristics} & \multirow{2}{*}{Solution Strategy} \\
\cline{3-7}
& & \makecell{Load-Dependent\\Energy Consumption} & \makecell{Obstacle\\Environment} & \makecell{Multiple\\Product Types}   & \makecell{Pickup\\Operations}  & \makecell{Locker\\Usage}  & \\
\hline
\cite{Jeong2019}        & FSTSP-ECNZ & $\surd$ & $\surd$ & & & & Heuristic \vspace{0.5mm}\\
\cite{Li2021}           & AC-2E-LRP & & & & $\surd$ & $\surd$ & Heuristic \vspace{0.5mm}\\
\cite{Boschetti2024}    & TSPDL & & & & & $\surd$ & Branch-and-cut \vspace{0.5mm}\\
\cite{Xiao2024}         & GVRPD-SR & $\surd$ & & & & & IALNS \vspace{0.5mm}\\
\cite{Meng2023}         & MDRP-PD & $\surd$ & & & $\surd$ & & Heuristic \vspace{0.5mm}\\
\cite{Liu2023}          & MDVRPSPDNF & & $\surd$ & & $\surd$ & & Heuristic \vspace{0.5mm}\\
\cite{Kong2024}          & C-TDRP & $\surd$ & $\surd$ & & & & DRL \vspace{0.5mm}\\
\cite{Zhu2024a}         & LDDMMP & & & & & $\surd$ & VNS-CPLEX \vspace{0.5mm}\\
\cite{DellAmico2023}    & PDVRPLTW & & & & $\surd$ & $\surd$ & Branch-and-cut \vspace{0.5mm}\\
This paper & LTDRP-PDNF    & $\surd$ & $\surd$ & $\surd$ & $\surd$ & $\surd$ & DRL \vspace{0.5mm}\\
\hline
\end{tabular*}
\end{table*}

% C. Deep reinforcement learning for routing problems
\subsection{Deep reinforcement learning for routing problems}\label{sec:Literature_DRL}

Over the past few years, there's been a growing consideration of DRL as an effective strategy for the vehicle routing problem (VRP) \cite{Che2024,Ren2022}, where the decision of the next node to visit is governed by a learned policy (Table \ref{table1}). \cite{Kool2018} designed a graph attention-based deep neural network using Google's Transformer architecture, training it through reinforcement learning to demonstrate its efficacy in solving the VRP. \cite{Li2022a} proposed a DRL approach employing a heterogeneous attention mechanism to solve a VRP variant known as the pickup and delivery problem, characterized by the inherent precedence constraint that pickup nodes must be visited before their corresponding delivery nodes. To address the multiobjective vehicle routing problem with time windows, \cite{Wu2025} proposed a weight-aware DRL approach to generate initial solutions, followed by optimization with the Non-dominated Sorting Genetic Algorithm II. \cite{Wang2024} proposed a neural heuristic method based on DRL to address both the classical and extended versions of the vehicle routing problem with backhauls (VRPB). The study modeled the VRPB using a graph representation and formulated its solution construction process as an MDP.

Given the effectiveness of DRL in addressing the combinatorial nature of VRP action spaces \cite{Zhu2024c,Bogyrbayeva2024} several studies have explored its application to the truck–drone routing problem. \cite{Bogyrbayeva2023} proposed a DRL approach that integrates an end-to-end model with long short-term memory networks to solve the traveling salesman problem with drone (TSP-D), in which a single truck equipped with a single drone is responsible for delivering customer orders.  \cite{Liu2022c} extended the FSTSP by incorporating stochastic travel times, formulating the FSTSP-STT, which considers the variability of travel times within the road network. After formulating the problem as an MDP, the study applied DRL algorithms, including deep Q-Network and advantage actor-critic, to solve the model. 
\cite{Kong2024} employed a DRL-based approach using the pointer network to optimize the collaborative truck–drone routing problem (C-TDRP). This problem involves launching a fleet of drones from trucks to serve randomly distributed customers, aiming to minimize delivery costs.  \cite{Peng2025} addressed a cooperative path-planning problem for trucks and drones, specifically applied to humanitarian aid distribution. The problem was modeled as a Markov game, and a multi-agent DRL algorithm was developed to obtain optimal routing strategies.

\section{Problem Description}

\begin{figure}[!t]
\centering
\includegraphics[width=3.5in]{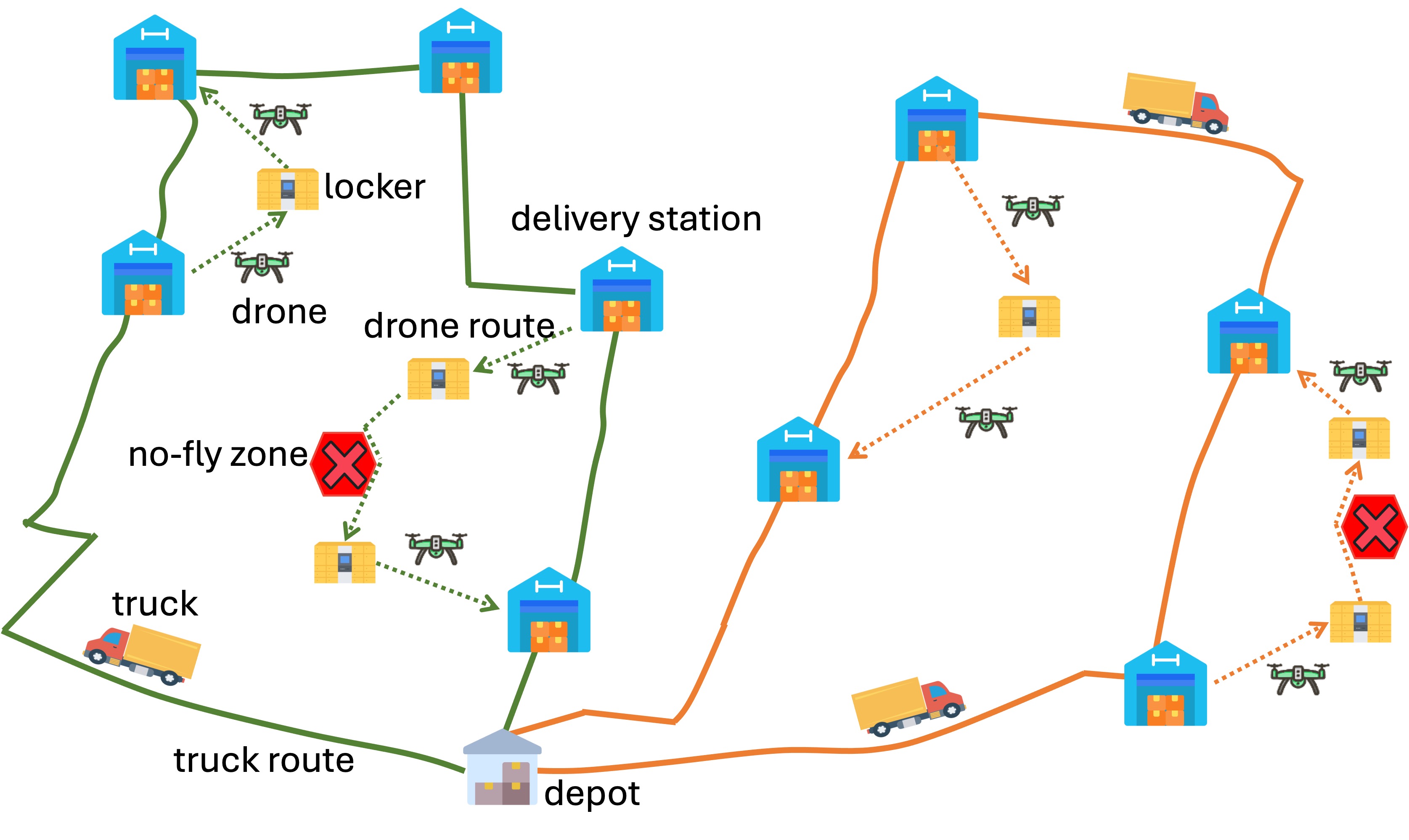}
\caption{A sketch for LTDRP-PDNF. }
\label{fig1}
\end{figure}

LTDRP-PDNF is defined as follows: A fleet of identical delivery trucks, each equipped with a drone, departs simultaneously from a central depot to serve delivery stations and lockers within a specified area. When a truck arrives at a station, its carried drone is deployed to deliver packages to nearby lockers. After launching the drone, the truck proceeds to the next delivery station. Each locker has both delivery and pickup demands for multiple product types, with each type possessing a distinct weight. Within the constraints of payload capacity and energy consumption, a drone can serve multiple lockers in a single trip. If a planned delivery route intersects a no-fly zone, the drone is required to take an alternative path that bypasses the restricted area. Drone battery replacement can occur at both trucks and lockers. When a drone is unable to continue serving the next customer, it must proceed to the nearest delivery station location to rendezvous with the truck and unload its cargo. After a drone rendezvous with its designated truck, it will depart from the truck fully loaded to continue the delivery tasks. Once all services for delivery stations and lockers are completed, the drone can either return to the depot with its truck or fly back independently. Fig. \ref{fig1} presents an illustrative example of LTDRP-PDNF.
The objective of LTDRP-PDNF is to fulfill all demands at the lowest total cost, and several key assumptions for this problem are detailed below: 

\begin{enumerate}[
    label=(\roman*)
]
\item{Lockers are assumed to have unlimited parcel capacity, thereby disregarding internal packing constraints \cite{Zhu2024b}.}
\item{The drone carried by each truck can only be launched and recovered from that truck.}
\item{Drones may be unloaded onto their carrying truck but are assumed to be fully loaded with delivery parcels before each takeoff. A single drone is capable of visiting multiple lockers during each flight, subject to its load and battery capacity limits.}
\item{The drone's power consumption is affected by its payload. Once the drone departs from current node (truck or locker after service), its battery will be swapped with a fully charged one.}
\item{Both trucks and drones maintain constant travel speeds, with drones always flying faster than trucks. A drone can only be
launched once the truck has arrived at the designated launch point.}
\item{The time spent on launching, landing, and servicing at each location is disregarded.}
\end{enumerate}

\section{Markov Decision Process Model}

The solution construction process of LTDRP-PDNF can be naturally viewed as a sequential decision-making problem. At each decision epoch, the selected truck or drone service affects subsequent routing decisions through changes in vehicle locations, residual capacities, locker demands, and truck--drone synchronization states. To capture these dynamic interactions and facilitate policy learning, LTDRP-PDNF is formulated as an MDP. The MDP is defined by the state space, action space, transition dynamics, and reward function described below. Table \ref{table2} summarizes the notations used in this work.

\begin{table}[!t]
\caption{Notations}
\label{table2}
\centering
\begin{tabular}{lp{6cm}}
\hline
Notations & Description \\
\hline
$N_{tr}$ & Set of distribution stations \\
$N_{dr}$ & Set of lockers \\
$M$ & Set of trucks \\
$K$ & Set of drones \\
$B$ & Set of product types, $B=\{b_1,b_2,\dots,b_n\}$ \\
$\tau_i^t$ & Delivery demand of distribution station $i$ at epoch $t$ \\
$d_{j,b_l}^{t}$ & Delivery demand of locker $j$ for product type $b_l$ at epoch $t$ \\
$p_{j,b_l}^{t}$ & Pickup demand of locker $j$ for product type $b_l$ at epoch $t$ \\
$\lambda_{b_l}$ & Unit weight of product type $b_l$ \\
$c_{tr,m}^{t}$ & Current destination node of truck $m$ at epoch $t$ \\
$q_{tr,m}^{t}$ & Remaining load of truck $m$ at epoch $t$  \\
$U_{tr,m}^{t}$ & Number of service legs completed by truck $m$ up to epoch $t$ \\
$o_{tr,m}^{t}$ & Remaining time for truck $m$ to arrive at its current destination \\
$v_{tr,m}^{t}$ & Set of distribution stations visited by truck $m$ up to epoch $t$ \\
$c_{dr,k}^{t}$ & Current destination node of drone $k$ at epoch $t$ \\
$q_{dr,k,b_l}^{t}$ & Remaining quantity of product type $b_l$ carried by drone $k$ for delivery at epoch $t$ \\
$q_{dr,k}^{t}$ & Vector of remaining delivery quantities carried by drone $k$, i.e., $q_{dr,k}^{t}=\{q_{dr,k,b_l}^{t}\}_{l=1}^{n}$ \\
$w_{dr,k}^{t}$ & Total payload currently carried by drone $k$ at epoch $t$ \\
$\beta _{dr,k}^{t}$ & Remaining battery level of drone $k$ at epoch $t$ \\
$U_{dr,k}^{t}$ & Number of service legs completed by drone $k$ up to epoch $t$ \\
$o_{dr,k}^{t}$ & Remaining time for drone $k$ to arrive at its current destination \\
$v_{dr,k}^{t}$ & Set of lockers visited by drone $k$ up to epoch $t$ \\
$a_{tr,m}^{t}$ & Next target node selected for truck $m$ at epoch $t$ \\
$a_{dr,k}^{t}$ & Next target node selected for drone $k$ at epoch $t$ \\
$e_{i,j}^{tr}$ & Travel time from node $i$ to node $j$ by truck \\
$e_{i,j}^{dr}$ & Travel time from node $i$ to node $j$ by drone \\
$r^{tr}(i,j)$ & Travel distance from node $i$ to node $j$ by truck \\
$r^{dr}(i,j)$ & Actual flight distance from node $i$ to node $j$ by drone, including no-fly-zone detours if needed \\
$V$ & Set of vertices of the no-fly-zone, arranged in a clockwise order \\
$F_k$ & Binary indicator equal to 1 if the flight path of drone $k$ crosses a no-fly-zone, and 0 otherwise \\
$\alpha $ & Battery energy consumption coefficient \\
$w_{dr,s}$ & Curb weight of each drone \\
$Q_{tr}$ & Truck capacity \\
$\beta_{\max}$ & Maximum battery capacity of a drone \\
$Q_{dr}$ & Maximum payload capacity of a drone \\
$q_{dr,k}^{0}$ & Replenished delivery-quantity vector of drone $k$ before a new mission \\
$w_{dr,k}^{0}$ & Initial payload of drone $k$ after replenishment before a new mission \\
$z_1, z_2$ & Unit transportation costs for trucks and drones \\
$s_1, s_2$ & Per-use costs for trucks and drones \\
\hline
\end{tabular}
\end{table}

\subsection{Decision Epoch}

The truck and drone operate as two types of decision-making agents in LTDRP-PDNF. Since truck movements and drone operations are not necessarily synchronized, we adopt an event-driven decision mechanism. A new decision epoch is triggered whenever a truck arrives at a distribution station, a drone arrives at a locker, or a drone returns to its associated truck. This event-driven representation avoids unnecessary time discretization and allows the MDP to explicitly capture the asynchronous coordination between trucks and drones.

\subsection{State Domain}

The global state at epoch $t$ is denoted as $S_t=(M_t,D_t,L_t,DS_t)$, where $M_t$, $D_t$, $L_t$, and $DS_t$ represent the truck state, drone state, locker state, and distribution-station state, respectively. The truck state is defined as $M_t=\{(c_{tr,m}^{t},q_{tr,m}^{t},U_{tr,m}^{t},o_{tr,m}^{t},v_{tr,m}^{t})\}_{m\in M}$. It includes the current destination node, remaining load, usage count, remaining travel time, and visited distribution stations of each truck. The drone state is defined as $D_t=\{(c_{dr,k}^{t},q_{dr,k}^{t},w_{dr,k}^{t},\beta_{dr,k}^{t},U_{dr,k}^{t},o_{dr,k}^{t},v_{dr,k}^{t})\}_{k\in K}$. It records the current destination node, remaining delivery quantities by product type, total payload, remaining battery level, usage count, remaining travel time, and visited lockers of each drone. If a truck or drone is traveling between two nodes at epoch $t$, its current node variable $c_{tr,m}^{t}$ or $c_{dr,k}^{t}$ is set to the destination node of the ongoing movement. If it is located exactly at a node, the corresponding variable is set to the index of that node.

The locker and distribution-station states are represented as $L_t=\{(d_{j,b_l}^{t},p_{j,b_l}^{t})\mid j\in N_{dr}, b_l\in B\}$ and $DS_t=\{\tau_i^t\mid i\in N_{tr}\}$. A distribution station only has delivery demand, whereas a locker may have both delivery and pickup demands for multiple product types. At the initial epoch, trucks and drones are fully replenished, i.e., $q_{tr,m}^{0}=Q_{tr}$, $q_{dr,k}^{0}$ is the initial delivery-quantity vector satisfying $w_{dr,k}^{0}\leq Q_{dr}$, and $\beta_{dr,k}^{0}=\beta_{\max}$.

\subsection{Action Domain}

At each decision epoch, an available truck or drone selects its next target node. The joint action is denoted as $a_t=(\{a_{tr,m}^{t}\}_{m\in M},\{a_{dr,k}^{t}\}_{k\in K})$. For truck $m$, the feasible target set is $a_{tr,m}^{t}\in N_{tr}\setminus v_{tr,m}^{t}$, where $a_{tr,m}^{t}$ denotes the next distribution station to be visited by the truck. For drone $k$, the feasible target set is $a_{dr,k}^{t}\in N_{dr}\setminus v_{dr,k}^{t}$, where $a_{dr,k}^{t}$ denotes the next locker to be served by the drone. Infeasible actions that violate capacity, battery, or service-completion constraints are excluded by the action mask during policy generation.

Given the geometric principle that a regular hexagon maximizes the area for a polygon with a given perimeter \cite{Sun2024}, we use regular hexagonal cells to represent restricted airspace. For a drone flight leg from the current node $c_{dr,k}^{t}$ to the next target node $a_{dr,k}^{t}$, let $p_s$ and $p_e$ denote the coordinates of the start and end points, respectively. Let $F_k$ be a binary indicator that equals 1 if the straight segment from $p_s$ to $p_e$ intersects a no-fly-zone, and 0 otherwise. When $F_k=1$, let $v_s$ and $v_e$ denote the boundary vertices of the corresponding hexagonal no-fly-zone that are closest to $p_s$ and $p_e$, respectively. The vertices of the hexagon are indexed along the polygon boundary, and $\mathcal{E}_{s,e}^{cw}$ and $\mathcal{E}_{s,e}^{ccw}$ denote the ordered sets of boundary edges from $v_s$ to $v_e$ in the clockwise and counterclockwise directions, respectively. The corresponding detour distances are calculated as follows:
\begin{equation}
\label{eq_MDP_1}
r_{cw} = \|p_s-v_s\| + \sum_{(i,j) \in \mathcal{E}_{s,e}^{cw}} \|v_i-v_j\| + \|v_e-p_e\|
\end{equation}
%%%
\begin{equation}
\label{eq_MDP_2}
r_{ccw} = \|p_s-v_s\| + \sum_{(i,j) \in \mathcal{E}_{s,e}^{ccw}} \|v_i-v_j\| + \|v_e-p_e\|
\end{equation}
%%%
\begin{equation}
\label{eq_MDP_3}
r^{dr}(c_{dr,k}^{t},a_{dr,k}^{t}) =
\begin{cases}
\min\{r_{cw},r_{ccw}\}, & F_k=1,\\
\|p_s-p_e\|, & F_k=0.
\end{cases}
\end{equation}
%%%
\begin{equation}
\label{eq_MDP_4}
\Delta\beta_{dr,k}^{t}
=
\alpha\left(w_{dr,s}+w_{dr,k}^{t}\right)
r^{dr}(c_{dr,k}^{t},a_{dr,k}^{t})
\end{equation}
%%%
Equations (\ref{eq_MDP_1}) and (\ref{eq_MDP_2}) represent the clockwise and counterclockwise detour distances around a no-fly-zone. Equation (\ref{eq_MDP_3}) calculates the actual flight distance from the current node to the next target node. If the straight flight segment intersects a no-fly-zone, the drone selects the shorter of the two feasible detour paths. Otherwise, the Euclidean distance between the two nodes is used. Equation (\ref{eq_MDP_4}) calculates the energy consumption of drone $k$ for the selected flight leg \cite{Meng2023}.

\subsection{State Transition Function}

Let $\Delta t$ denote the time elapsed from epoch $t$ to epoch $t+1$, which is determined by the earliest completion time among the ongoing truck and drone movements. For the truck and drone selected at epoch $t$, the elapsed time can be written as
$
\Delta t
=
\min\left\{
o_{tr,m}^{t}+e_{c_{tr,m}^{t},a_{tr,m}^{t}}^{tr},
\;
o_{dr,k}^{t}+e_{c_{dr,k}^{t},a_{dr,k}^{t}}^{dr}
\right\}.
$
This event-driven update reduces the remaining travel times of all ongoing movements and triggers the next decision epoch once at least one vehicle completes its current service leg.

\subsubsection{Truck-related state transition}

Truck-related state transitions include the state changes of both trucks and distribution stations. For any truck $m\in M$, the state transition is given as follows:
%%%
\begin{equation}
\label{eq_MDP_5}
c_{tr,m}^{t+1}=a_{tr,m}^{t}
\end{equation}
%%%
\begin{equation}
\label{eq_MDP_6}
q_{tr,m}^{t+1}
=
q_{tr,m}^{t}-\tau_{a_{tr,m}^{t}}^{t}
\end{equation}
%%%
\begin{equation}
\label{eq_MDP_7}
U_{tr,m}^{t+1}=U_{tr,m}^{t}+1
\end{equation}
%%%
\begin{equation}
\label{eq_MDP_8}
o_{tr,m}^{t+1}
=
o_{tr,m}^{t}
+
e_{c_{tr,m}^{t},a_{tr,m}^{t}}^{tr}
-
\Delta t
\end{equation}
%%%
\begin{equation}
\label{eq_MDP_9}
v_{tr,m}^{t+1}=v_{tr,m}^{t}\cup \{a_{tr,m}^{t}\}
\end{equation}
%%%
\begin{equation}
\label{eq_MDP_10}
\tau_{a_{tr,m}^{t}}^{t+1}=0
\end{equation}
%%%
Equation (\ref{eq_MDP_5}) updates the current destination of truck $m$. Equation (\ref{eq_MDP_6}) updates the remaining truck load after serving the selected distribution station. Equation (\ref{eq_MDP_7}) updates the truck usage count. Equation (\ref{eq_MDP_8}) updates the remaining travel time under the event-driven time advance $\Delta t$. Equation (\ref{eq_MDP_9}) updates the set of visited distribution stations. Equation (\ref{eq_MDP_10}) sets the demand of the served distribution station to zero for the next epoch.

\subsubsection{Drone-related state transition}

Drone-related state transitions include the state changes of both drones and lockers. For drone $k$, the selected locker $a_{dr,k}^{t}$ can be successfully served only when the following conditions are simultaneously satisfied: the required flight energy does not exceed the maximum battery capacity, the drone carries sufficient products to meet the delivery demand of the selected locker, and the payload after completing both delivery and pickup services does not exceed the drone payload capacity. Specifically, these feasibility conditions are expressed as
$\Delta \beta_{dr,k}^{t} \leq \beta_{\max}$,
$q_{dr,k,b_l}^{t}-d_{a_{dr,k}^{t},b_l}^{t}\geq 0$ for all $b_l\in B$, and
$\sum_{b_l\in B}\lambda_{b_l}
(q_{dr,k,b_l}^{t}-d_{a_{dr,k}^{t},b_l}^{t}+p_{a_{dr,k}^{t},b_l}^{t})\leq Q_{dr}$.
If any of these conditions is violated, the drone returns to the truck for unloading and replenishment before starting a new mission.
%%%
\begin{equation}
\label{eq_MDP_11}
q_{dr,k}^{t+1} =
\begin{cases}
\left\{ q_{dr,k,b_l}^{t}-d_{a_{dr,k}^{t},b_l}^{t} \right\}_{b_l\in B}, & \begin{array}{@{}l@{}} \text{if locker } a_{dr,k}^{t} \\ \text{is feasibly served,} \end{array} \\[1.2em]
q_{dr,k}^{0}, & \text{otherwise.}
\end{cases}
\end{equation}
%%%
\begin{equation}
\label{eq_MDP_12}
w_{dr,k}^{t+1} =
\begin{cases}
\begin{aligned}
&\sum_{b_l\in B}\lambda_{b_l} \big( q_{dr,k,b_l}^{t} - d_{a_{dr,k}^{t},b_l}^{t} \\
&\qquad\qquad + p_{a_{dr,k}^{t},b_l}^{t} \big),
\end{aligned} & \begin{array}{@{}l@{}} \text{if locker } a_{dr,k}^{t} \\ \text{is feasibly served,} \end{array} \\[1.8em]
Q_{dr}, & \text{otherwise.}
\end{cases}
\end{equation}
%%%

\begin{equation}
\label{eq_MDP_13}
\beta_{dr,k}^{t+1}=\beta_{\max}
\end{equation}
%%%
\begin{equation}
\label{eq_MDP_14}
U_{dr,k}^{t+1}=U_{dr,k}^{t}+1
\end{equation}
%%%
\begin{equation}
\label{eq_MDP_15}
o_{dr,k}^{t+1}
=
o_{dr,k}^{t}
+
e_{c_{dr,k}^{t},a_{dr,k}^{t}}^{dr}
-
\Delta t
\end{equation}
%%%
\begin{equation}
\label{eq_MDP_16}
v_{dr,k}^{t+1}=v_{dr,k}^{t}\cup \{a_{dr,k}^{t}\}
\end{equation}

For each product type $b_l\in B$, the delivery and pickup demands of the selected locker are updated as follows.
%%%
\begin{equation}
\label{eq_MDP_17}
d_{a_{dr,k}^{t},b_l}^{t+1} =
\begin{cases}
0, & \text{if the service is completed}, \\[0.5em]
d_{a_{dr,k}^{t},b_l}^{t}, & \text{otherwise},
\end{cases}
\end{equation}
%%%
\begin{equation}
\label{eq_MDP_18}
p_{a_{dr,k}^{t},b_l}^{t+1} =
\begin{cases}
0, & \text{if the service is completed}, \\[0.5em]
p_{a_{dr,k}^{t},b_l}^{t}, & \text{otherwise}.
\end{cases}
\end{equation}
%%% 
Equations (\ref{eq_MDP_11}) and (\ref{eq_MDP_12}) update the remaining delivery-quantity vector and total payload of drone $k$, respectively. When the selected locker is feasibly served, the delivery quantities carried by the drone are reduced by the corresponding locker delivery demands, and the total payload is updated by considering both the remaining delivery parcels and the newly picked-up parcels. If the battery, payload, or product-availability constraint is violated, the drone returns to the truck for unloading and replenishment before starting a new mission. Equation (\ref{eq_MDP_13}) indicates that the drone receives a fully charged battery before departure. Equation (\ref{eq_MDP_14}) updates the drone usage frequency. Equations (\ref{eq_MDP_15}) and (\ref{eq_MDP_16}) update the remaining travel time and the set of visited lockers, respectively. Equations (\ref{eq_MDP_17}) and (\ref{eq_MDP_18}) update the delivery and pickup demands of the selected locker for each product type. The corresponding demands are set to zero only after the locker service is completed; otherwise, they remain unchanged for subsequent service.

\subsection{Reward}

The immediate cost at epoch $t$ consists of transportation cost and vehicle-use cost:
%%%
\begin{equation}
\label{eq_MDP_19}
C^{t}=\Delta C_1^{t}+\Delta C_2^{t}
\end{equation}
%%%
\begin{equation}
\label{eq_MDP_20}
\Delta C_1^{t}
=
\sum_{m\in M}
z_1 r^{tr}(c_{tr,m}^{t},a_{tr,m}^{t})
+
\sum_{k\in K}
z_2 r^{dr}(c_{dr,k}^{t},a_{dr,k}^{t})
\end{equation}
%%%
\begin{equation}
\label{eq_MDP_21}
\Delta C_2^{t}
=
\sum_{m\in M}
s_1\left(U_{tr,m}^{t+1}-U_{tr,m}^{t}\right)
+
\sum_{k\in K}
s_2\left(U_{dr,k}^{t+1}-U_{dr,k}^{t}\right)
\end{equation}
%%%
Equation (\ref{eq_MDP_19}) defines the immediate operational cost at epoch $t$. Equation (\ref{eq_MDP_20}) calculates the transportation cost of truck and drone movements, and Equation (\ref{eq_MDP_21}) calculates the incremental usage cost of trucks and drones. Since the objective is to minimize the total operational cost, the immediate reward is defined as the negative cost:
$
R(S_t,a_t)=-C^{t}.
$

\section{Solution Method}

\begin{figure}[!t]
\centering
\includegraphics[width=\columnwidth]{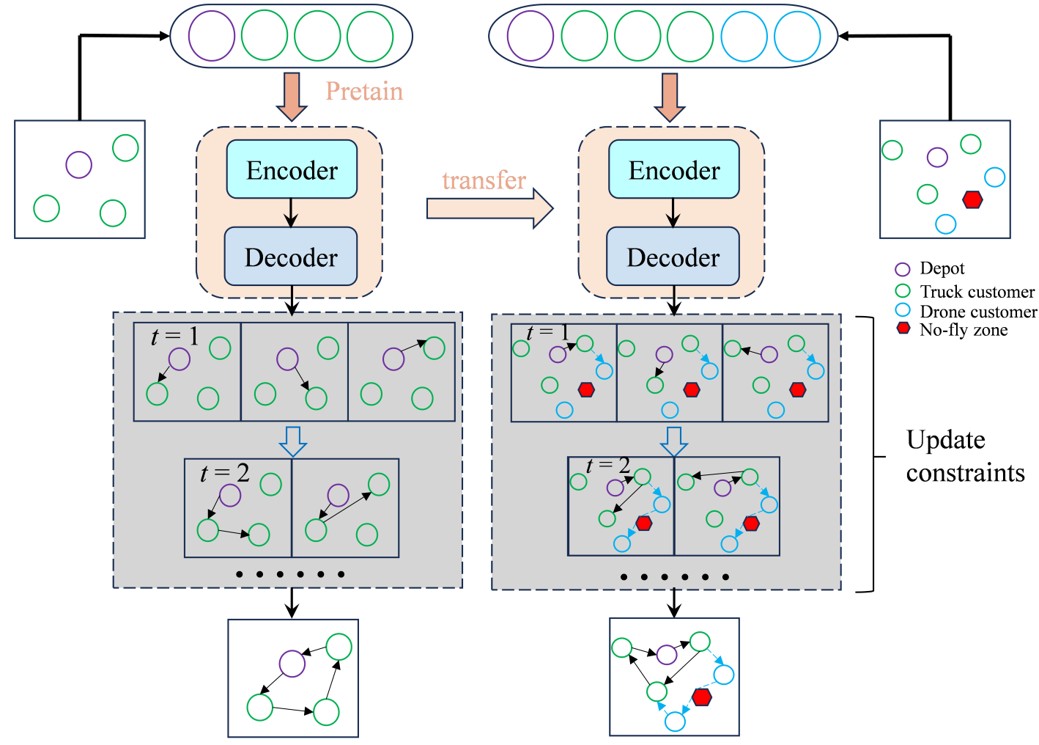}
\caption{Overall structure}
\label{fig2}
\end{figure}

In this section, we provide a detailed description of the proposed two-stage DRL-based neural heuristic. In the first stage, we employ an end-to-end model that uses an attention-based encoder for adaptive feature extraction and a Bi-GRU network decoder for sequential decision-making to solve the CVRP with truck-only delivery. In the second stage, we introduce a transfer strategy that leverages the solution parameters learned in the first stage and employs a hybrid dispatch assignment heuristic that constructs drone delivery routes, thereby transferring solutions from the CVRP to LTDRP-PDNF. An overview of its architecture is illustrated in Fig. \ref{fig2}.

\subsection{Stage I: Solving the CVRP}

\begin{figure}[!t]
\centering
\includegraphics[width=\columnwidth]{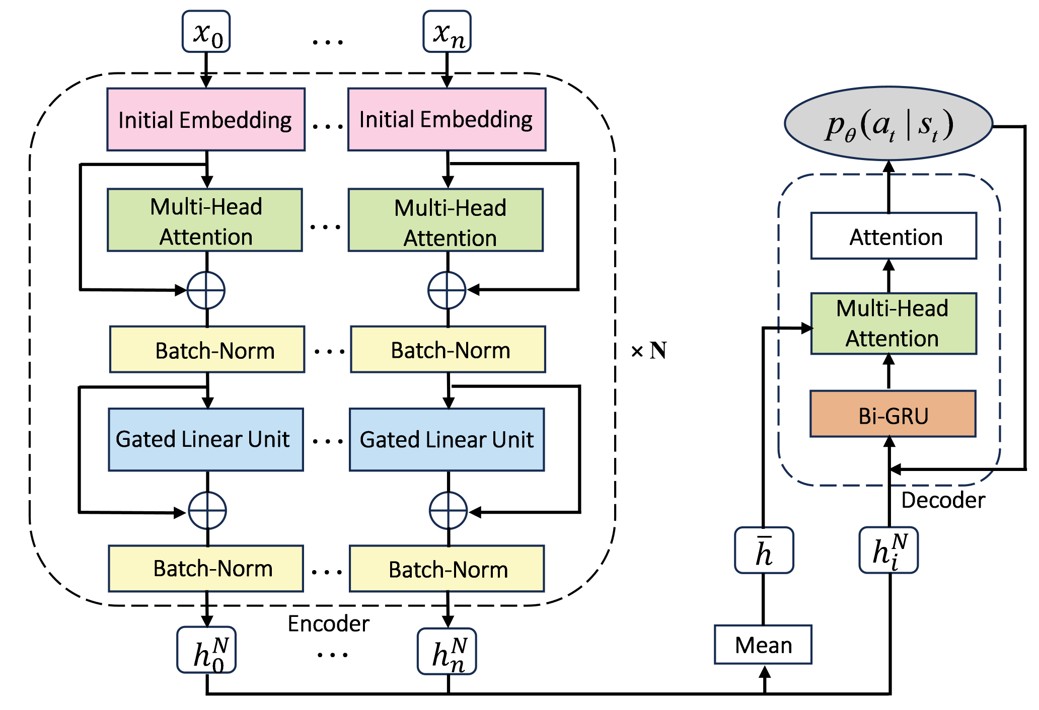}
\caption{The encoder-decoder structure for solving CVRP}
\label{fig3}
\end{figure}

The first stage aims to construct a truck-only delivery plan that provides the routing backbone for the subsequent locker-based truck--drone coordination problem. This subproblem is formulated as a CVRP, where trucks depart from the depot, serve customer nodes under vehicle capacity constraints, and return to the depot when necessary. We design an encoder--decoder neural policy to generate truck routes sequentially. The encoder maps the coordinate and demand information of depot and customer nodes into latent representations, while the decoder selects the next node to visit according to the current routing state and feasibility constraints. To better capture sequential routing information, a bidirectional gated recurrent unit (Bi-GRU) is embedded in the decoder. The forward GRU encodes the already constructed route context, whereas the backward GRU provides complementary information from the candidate-node sequence. The overall architecture is shown in Fig. \ref{fig3}.

\subsubsection{Encoder}

Let $\mathcal{V}=\{0,1,\dots,n\}$ denote the node set, where node $0$ represents the depot and $\mathcal{C}=\{1,2,\dots,n\}$ represents the customer set. Each node is described by its coordinate vector $x_i$. For customer nodes, the demand is further normalized by the truck capacity $P$, i.e., $\delta_i=d_i/P$. The initial representation of node $i$ is obtained through a trainable embedding layer:
\begin{equation}
\label{eq1}
z_i^{(0)} =
\begin{cases}
A_0 x_i+a_0, & i=0,\\
A_c[x_i;\delta_i]+a_c, & i\in\mathcal{C}.
\end{cases}
\end{equation}
where $z_i^{(0)}$ denotes the initial node embedding, $[\cdot;\cdot]$ is the concatenation operator, and $A_0$, $a_0$, $A_c$, and $a_c$ are trainable parameters.

The depot and customer nodes contain different types of information. The depot is represented only by its spatial position, whereas each customer node contains both spatial information and demand information. To capture interactions among these heterogeneous nodes, we employ a multi-head attention module in the encoder. Since the CVRP is defined on a complete routing graph, each node can exchange information with all other nodes during encoding. For the $r$-th attention head in layer $\ell$, the transformed representations are defined as:
\begin{equation}
\label{eq2}
Q_{i}^{\ell,r}=W_{Q}^{\ell,r}z_{i}^{(\ell-1)},\;
K_{i}^{\ell,r}=W_{K}^{\ell,r}z_{i}^{(\ell-1)},\;
V_{i}^{\ell,r}=W_{V}^{\ell,r}z_{i}^{(\ell-1)}
\end{equation}
%%%
where $W_{Q}^{\ell,r}$, $W_{K}^{\ell,r}$, and $W_{V}^{\ell,r}$ are head-specific trainable projection matrices.

The attention compatibility between nodes $i$ and $j$ is calculated by a scaled inner product:
\begin{equation}
\label{eq3}
\eta_{ij}^{\ell,r}=
\frac{\left(Q_{i}^{\ell,r}\right)^{\top}K_{j}^{\ell,r}}
{\sqrt{d_r}}
\end{equation}
where $d_r$ denotes the dimension of the projected representation in each attention head.
%%%
\begin{equation}
\label{eq4}
\omega_{ij}^{\ell,r}
=
\frac{\exp\left(\eta_{ij}^{\ell,r}\right)}
{\sum_{u\in\mathcal{V}}\exp\left(\eta_{iu}^{\ell,r}\right)}
\end{equation}
%%%

The representation aggregated by the $r$-th head is obtained as:
%%%
\begin{equation}
\label{eq5}
s_{i}^{\ell,r}
=
\sum_{j\in\mathcal{V}}\omega_{ij}^{\ell,r}V_{j}^{\ell,r}
\end{equation}

The outputs of all attention heads are concatenated and projected to generate the encoder message for node $i$:
\begin{equation}
\label{eq6}
m_{i}^{\ell}
=
W_{O}^{\ell}
\left[
s_{i}^{\ell,1};
s_{i}^{\ell,2};
\dots;
s_{i}^{\ell,H}
\right]
\end{equation}
where $H$ is the number of attention heads and $W_{O}^{\ell}$ is a trainable output projection matrix. In this study, $H$ is set to 8.

A residual connection and normalization operation are then applied to stabilize the representation update:
\begin{equation}
\label{eq7}
\widehat{z}_{i}^{\ell}
=
\operatorname{Norm}^{\ell}
\left(
z_{i}^{(\ell-1)}+m_{i}^{\ell}
\right)
\end{equation}

To enhance the nonlinear transformation capability of the encoder, we use a gated linear unit (GLU) instead of a conventional fully connected feed-forward layer. Following \cite{Shazeer2020}, the feed-forward block is written as:
\begin{equation}
\label{eq8}
\operatorname{FF}^{\ell}(\widehat{z}_{i}^{\ell})
=
\left(
W_{g}^{\ell}\widehat{z}_{i}^{\ell}+b_{g}^{\ell}
\right)
\otimes
\operatorname{GELU}
\left(
W_{f}^{\ell}\widehat{z}_{i}^{\ell}+b_{f}^{\ell}
\right)
\end{equation}
where $\otimes$ denotes the element-wise product, and $\operatorname{GELU}(\cdot)$ is the activation function.

The output of the feed-forward block is connected with its input through another residual path:
\begin{equation}
\label{eq9}
z_{i}^{\ell}
=
\operatorname{Norm}^{\ell}
\left(
\widehat{z}_{i}^{\ell}
+
\operatorname{FF}^{\ell}(\widehat{z}_{i}^{\ell})
\right)
\end{equation}

After the final encoder layer, a graph-level representation is obtained by averaging all node embeddings:
\begin{equation}
\label{eq10}
g
=
\frac{1}{|\mathcal{V}|}
\sum_{i\in\mathcal{V}}
z_{i}^{(N_e)}
\end{equation}
where $N_e$ is the number of encoder layers and $g$ represents the global embedding of the CVRP instance.

\subsubsection{Decoder}

The decoder constructs truck routes step by step. At each decision step $t$, the decoder observes the current truck location, the remaining vehicle capacity, the set of unserved customers, and the encoded node representations. To incorporate both forward and backward sequential context, the encoder outputs are processed by a Bi-GRU. For a given candidate sequence, the forward and backward hidden states are computed as:
\begin{equation}
\label{eq11}
\overrightarrow{r}_{i}^{t}
=
\operatorname{GRU}_{f}
\left(
z_{i}^{(N_e)},
\overrightarrow{r}_{i-1}^{t}
\right)
\end{equation}
%%%
\begin{equation}
\label{eq12}
\overleftarrow{r}_{i}^{t}
=
\operatorname{GRU}_{b}
\left(
z_{i}^{(N_e)},
\overleftarrow{r}_{i+1}^{t}
\right)
\end{equation}
where $\overrightarrow{r}_{i}^{t}$ and $\overleftarrow{r}_{i}^{t}$ are the forward and backward hidden states associated with node $i$ at decision step $t$, respectively.

The contextual representation of candidate node $i$ is formed by concatenating the two hidden states:
\begin{equation}
\label{eq13}
r_{i}^{t}
=
\left[
\overrightarrow{r}_{i}^{t};
\overleftarrow{r}_{i}^{t}
\right]
\end{equation}

Let $\ell_t$ denote the current location of the truck and $Q_t^{\mathrm{rem}}$ denote the remaining vehicle capacity at step $t$. The decoder context is defined as:
\begin{equation}
\label{eq14}
c_{t}
=
W_{c}
\left[
g;
z_{\ell_t}^{(N_e)};
Q_t^{\mathrm{rem}}/P
\right]
+
b_{c}
\end{equation}
where $W_{c}$ and $b_{c}$ are trainable parameters.

For each candidate node $i$, a routing feature vector is constructed as:
\begin{equation}
\label{eq15}
\phi_{t,i}
=
\left[
c_{\ell_t i};
d_i/P;
\mathbb{I}(i=0)
\right]
\end{equation}
where $c_{\ell_t i}$ is the travel cost from the current truck location $\ell_t$ to node $i$, and $\mathbb{I}(i=0)$ indicates whether the candidate node is the depot.

The selection score of node $i$ is computed by combining the decoder context, the candidate-node representation, and the routing feature vector:
\begin{equation}
\label{eq16}
\rho_{t,i}
=
w_{s}^{\top}
\tanh
\left(
W_{s}c_{t}
+
U_{s}r_{i}^{t}
+
B_{s}\phi_{t,i}
\right)
\end{equation}
where $w_{s}$, $W_{s}$, $U_{s}$, and $B_{s}$ are trainable parameters.

To ensure feasibility, the decoder masks invalid actions. Let $\mathcal{A}_{t}$ denote the feasible action set at step $t$, which includes unserved customers whose demands do not exceed the remaining vehicle capacity and the depot when route termination or vehicle reloading is allowed. The clipped and masked score is given by:
\begin{equation}
\label{eq17}
\bar{\rho}_{t,i}
=
\begin{cases}
\max\{-C,\min\{C,\rho_{t,i}\}\}, & i\in\mathcal{A}_{t},\\
-\infty, & i\notin\mathcal{A}_{t}.
\end{cases}
\end{equation}
where $C$ is a clipping threshold used to improve numerical stability.

The probability of selecting node $i$ as the next destination is obtained through softmax normalization:
\begin{equation}
\label{eq18}
p_{\theta}(a_t=i\mid s_t)
=
\frac{\exp(\bar{\rho}_{t,i})}
{\sum_{j\in\mathcal{A}_{t}}\exp(\bar{\rho}_{t,j})}
\end{equation}

The node with the largest selection probability is chosen as the next destination during greedy decoding. After the selection, the truck location, remaining capacity, served-customer set, and feasible action set are updated. This process is repeated until all customers are served and the truck returns to the depot.

\begin{algorithm}[!t]
\caption{Training Procedure}
\label{alg:training_procedure}
\begin{algorithmic}[1]
\renewcommand{\algorithmicrequire}{\textbf{Input:}}
\renewcommand{\algorithmicensure}{\textbf{Output:}}
\REQUIRE Node set $\mathcal{D}$, maximum number of epochs $E_{epochs}$, number of batches $B$, step limit $T$, instance size $N$, policy network $p_\theta$, early-stop threshold $k$
\ENSURE A policy network trained for efficient route generation
\STATE Randomly initialize policy parameters $\theta$;
\FOR{$epoch = 1$ \TO $E_{epochs}$}
    \STATE Generate $B$ random CVRP instances;
    \FOR{$batch = 1$ \TO $B$}
        \STATE Sample solution trajectories $\{\xi^1,\xi^2,\dots,\xi^N\}$ under the current policy;
        \FOR{$i = 1$ \TO $N$}
            \STATE Initialize $t \leftarrow 0$ and construct an empty route;
            \WHILE{$t < T$}
                \STATE Select action $a_t^i$ according to $p_\theta(a_t^i \mid s_t^i)$;
                \STATE Observe reward $r_t^i$ and update state $s_{t+1}^i$;
                \STATE $t \leftarrow t + 1$;
            \ENDWHILE
            \STATE Compute trajectory reward $\mathcal{R}(\xi^i)=\sum_{t=0}^{T}r_t^i$;
        \ENDFOR
        \STATE Update the baseline according to Eq. \eqref{eq19};
        \STATE Compute the policy-gradient estimate according to Eq. \eqref{eq20};
        \STATE Update parameters: $\theta \leftarrow \theta+\alpha\nabla_\theta J(\theta)$;
        \FOR{$i = 1$ \TO $N$}
            \STATE Apply 2-opt local search to trajectory $\xi^i$;
            \STATE $c_{fail} \leftarrow 0$;
            \WHILE{$c_{fail} < k$}
                \STATE Generate a new route by a 2-opt swap;
                \IF{the new route improves $\xi^i$}
                    \STATE Update $\xi^i$;
                    \STATE $c_{fail} \leftarrow 0$;
                \ELSE
                    \STATE $c_{fail} \leftarrow c_{fail}+1$;
                \ENDIF
            \ENDWHILE
        \ENDFOR
    \ENDFOR
\ENDFOR
\end{algorithmic}
\end{algorithm}

\subsubsection{Training}

The policy is trained using the REINFORCE algorithm. For each training instance with $D$ nodes, multiple complete trajectories are sampled under the current policy through Monte Carlo simulation. Let $\xi_i$ denote the $i$-th sampled trajectory and $R(\xi_i)$ denote its cumulative reward. Since the objective is to minimize the route cost, the reward is defined as the negative routing cost. To reduce the variance of the policy-gradient estimate, the average reward of the sampled trajectories is used as the baseline:
\begin{equation}
\label{eq19}
\bar{R}(s)
=
\frac{1}{D}
\sum_{i=1}^{D}
R(\xi_i)
\end{equation}

The policy parameters are updated by the Adam optimizer according to the following gradient estimate:
\begin{equation}
\label{eq20}
\nabla_{\theta}J(\theta)
\approx
\frac{1}{D}
\sum_{i=1}^{D}
\left(
R(\xi_i)-\bar{R}(s)
\right)
\nabla_{\theta}
\log p_{\theta}(\xi_i\mid s)
\end{equation}

After a complete truck route is generated, a lightweight 2-opt improvement procedure is applied to remove unnecessary edge crossings and further refine the solution. In each iteration, two non-adjacent edges are selected and reconnected in an alternative order. The new route replaces the current one only when it reduces the total routing cost. To avoid excessive local-search time, an early-stopping rule is adopted: if no improvement is found after several consecutive trials, the 2-opt procedure is terminated. The training and post-processing procedure is summarized in Algorithm \ref{alg:training_procedure}.

\subsection{Stage II: Solving LTDRP-PDNF}

After obtaining the solution parameters of the CVRP model, considering the similarity between CVRP and LTDRP-PDNF in terms of problem types, as well as the transferability of neural network models \cite{Neyshabur2020}, we design a transfer strategy to enable solution migration from the source domain (CVRP) to the target domain (LTDRP-PDNF). The implementation of the transfer strategy can be divided into three steps as follows.

\begin{enumerate}[
    label=(\roman*)
]
\item{The pre-trained DRL network is developed using extensive data from the source domain, to obtain an optimal set of neural network parameters.}
\item{The network parameters pretrained in the source domain are reused in the target domain to generate truck routes, significantly reducing the time required for model training.}
\item{A hybrid dispatch assignment heuristic is used to construct drone routes, enabling effective problem solving in the target domain.}
\end{enumerate}

The training parameters from the source domain are applied to generate truck routes in the target domain as follows:
\begin{equation}
\label{eq42}
\begin{cases}
\theta_t = \theta_s \\
C_t = \{C_s, C_{dr}\} \\
Y_t = F(C_t, \theta_t)
\end{cases}
\end{equation}
where $\theta_s, \theta_t$ represent the network parameters associated with the source and target domains, respectively; $C_s, C_t$ denote the input parameters of the source and target domains; $C_{dr}$ refers to the drone parameters and $Y_t$ represents the output of the target domain, i.e., the truck routes $R_{tr}$.

Since drone load capacity and battery capacity must be respected and checked throughout the pickup and delivery process, random methods may restrict the exploration capabilities of subsequent neighborhood searches. Moreover, payload and battery constraints need to be verified for each iteration, which increases computational overhead. To address this, we propose a hybrid dispatch assignment heuristic, refer to as MBP-ILS, which integrates a maximum battery and payload (MBP) method, responsible for generating initial drone routes, with an improved local search (ILS) algorithm, which then optimizes these routes. For a truck route $R \in R_{tr}$, Algorithm \ref{alg:drone_route} details a specific step in the generation process of an initial delivery route segment for its carried drone.

\begin{algorithm}[!t]
\caption{Drone Route Construction}
\label{alg:drone_route}
\begin{algorithmic}[1]
\renewcommand{\algorithmicrequire}{\textbf{Input:}}
\renewcommand{\algorithmicensure}{\textbf{Output:}}
\REQUIRE First delivery station $s_0$ visited by the truck; \\
Set of lockers $c_R$ assigned to $s_0$; \\
Delivery demand $d_i^l$ and pickup demand $p_i^l$ at locker $i$; \\
Number of product types $n$; \\
Drone location $h$, total load $w^t$, product load $w^l$; \\
Drone battery $\beta$, max battery $\beta_{\max}$; \\
Drone current destination $c_d$
\ENSURE Drone delivery routes $SI$
\STATE $SI \leftarrow [s_0]$;
\STATE Initialize $h, w^t, w^l, \beta \leftarrow s_0, Q_{dr}, \frac{Q_{dr}}{n}, \beta_{\max}$;
\WHILE{$c_R$ is not empty}
    \STATE $c_d = c_i \leftarrow$ Find the nearest locker;
    \IF{$w^l \ge d_i^l$ \textbf{and} $w^t \le Q_{dr}$ \textbf{and} $\alpha \cdot (w_s + w^t) \cdot r(h, c_i) \le \beta_{\max}$}
        \STATE $SI \leftarrow$ Insert $c_i$ to drone route;
        \STATE Update $h, w^t, w^l, \beta$:
        \STATE $h \leftarrow c_i$;
        \STATE $w^k \leftarrow w^l - d_i^l$;
        \STATE $w^t \leftarrow \sum_{l=1}^{n} (w^l + p_i^l)$;
        \STATE $\beta \leftarrow \beta_{\max}$;
        \STATE $c_R = c_R \setminus \{c_i\}$;
        \STATE $c_d = c_j \leftarrow$ Find the nearest locker;
        \IF{$w^l \ge d_j^l$ \textbf{and} $w^t \le Q_{dr}$ \textbf{and} $\alpha \cdot (w_s + w^t) \cdot r(h, c_j) \le \beta_{\max}$}
            \STATE $SI \leftarrow$ Insert $c_j$ to drone route;
            \STATE Update $h, w^t, w^l, \beta$:
            \STATE $h \leftarrow c_j$;
            \STATE $w^k \leftarrow w^l - d_j^l$;
            \STATE $w^t \leftarrow \sum_{k=1}^{n} (w^l + p_j^l)$;
            \STATE $\beta \leftarrow \beta_{\max}$;
            \STATE $c_R = c_R \setminus \{c_j\}$;
        \ELSE
            \STATE $c_d = s_1 \leftarrow$ Find the nearest delivery station (Truck has already arrived);
            \STATE $SI = $ Insert $s_1$ to drone route;
        \ENDIF
    \ELSE
        \STATE Nearest-neighbor reassign;
    \ENDIF
\ENDWHILE
\STATE \textbf{return} $SI$;
\end{algorithmic}
\end{algorithm}

Before constructing the initial routes, we employ a nearest-neighbor assignment and reassignment heuristic to allocate each locker. For $R \in R_{tr}$, the heuristic identifies the locker $c_i$ closest to each delivery station $s_i$ on the route, assigning each locker to its corresponding delivery station. After the initial assignment, any unassigned locker is reassigned to its nearest delivery station. This process continues until all lockers $N_{dr}$ are assigned to their respective routes.

Let $d_j^l, p_j^l$ be the delivery and pickup demand for product $l$ at locker $j$, respectively; $w_i^l$ be the drone's remaining payload capacity for product $l$ after serving locker $i$; $w_i^t$ be the total payload of the drone when departing from its current location to serve the next locker.

\begin{figure}[!t]
\centering
\includegraphics[width=\columnwidth]{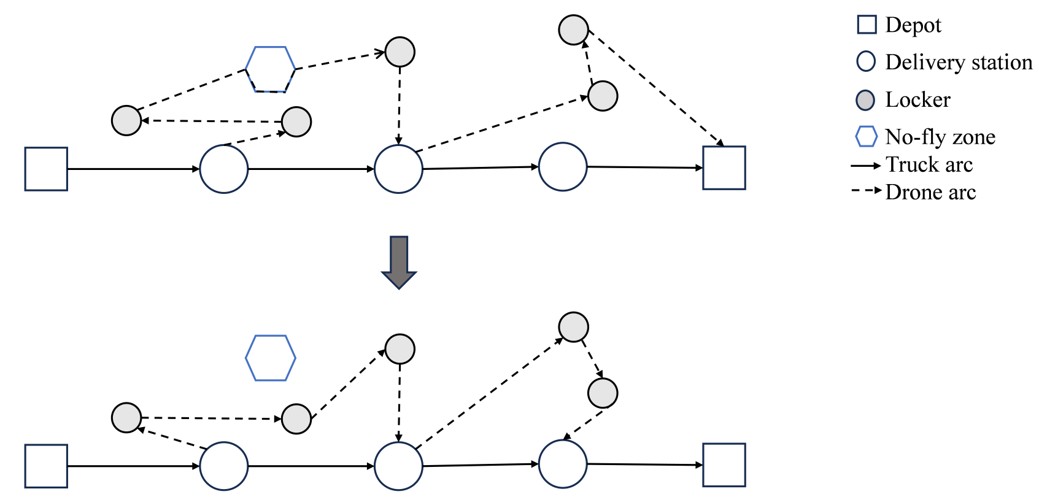}
\caption{Optimization process of the drone route}
\label{fig4}
\end{figure}

The MBP method can be described as follows: For $R \in R_{tr}$, the first delivery station $s_0$ visited on $R$ is selected as the starting point of the drone route. The lockers assigned to $s_0$ serve as the drone customers. A distance-based greedy algorithm is utilized to construct the initial drone route, starting from the locker $i$ that is closest to $s_0$. Subsequently, after servicing a locker $i$, the unvisited locker $j$ closest to $i$ is chosen as the next candidate service point. The locker $j$ is inserted into the drone route if both of the following constraints are satisfied during the delivery process. First, the payload constraint requires that the remaining payload capacity of the drone for each product type exceeds the delivery demand of the locker (i.e., $w_i^l \ge d_j^l$), and the total payload after servicing locker $j$ remains within the drone maximum load capacity (i.e., $w_j^t \le Q_{dr}$). Second, the battery constraint ensures that the energy consumption required for the drone to travel from its current location to the locker does not exceed its maximum battery capacity (i.e., $\alpha \cdot (w_s + w_i^t) \cdot r(i, j) \le \beta_{\max}$). If, after servicing the current locker $i$, the drone cannot proceed to the next candidate locker $j$ because inserting $j$ would violate either constraint the payload or battery constraint, the drone travels to its nearest delivery station where the truck has already arrived (i.e., $t_1 \le t_2$, where $t_1$ is drone arrival and $t_2$ is truck arrival at the delivery station), and a new drone route is initiated at the delivery station. This operation continues until all lockers have received service.
The tight constraints of the MBP method make its solutions susceptible to local optima. To further improve the initial solution, an improved local search (ILS) algorithm is utilized. We employ a 2-opt operation to swap the access order of two points, which helps reduce drone travel distances and paths intersecting no-fly zones. Moreover, in determining the final rendezvous point, we prioritize the solution with the lowest cost rather than simply its accessibility. The optimization process is depicted in Fig.~\ref{fig4}.

\section{Computational Experiments}

\subsection{Experimental Settings}

\subsubsection{Datasets}

Our experiments primarily focus on two types of problem datasets, i.e., CVRP and LTDRP-PDNF. Regarding the former dataset, we use the benchmark datasets employed by \cite{Kool2018,Kwon2020}, which include random instances with customer node sizes of 20, 50, and 100, each containing 10,000 instances. Specifically, the origin is set as the depot node, and the coordinates of the customer nodes are generated by uniformly sampling from the $[0,1] \times [0,1]$ region. The customer demand is generated by uniformly sampling from the range $[1,10]$. Depending on the node size, the truck capacities are set to 30, 40, and 50 kg, with a unit transportation cost of \$1. The objective is to service all customer nodes using the shortest possible route. For more details on the dataset generation method, refer to \url{https://github.com/wouterkool/attention-learn-to-route}.

Regarding the latter dataset, given the absence of publicly available benchmarks for the proposed LTDRP-PDNF, we generated synthetic instances by adapting established instances that can be found online at the VRP with Pick-up and Deliveries (VRP Web (dorronsoro.es)). The dataset includes instances with 20, 50, and 100 customer nodes, with 100 instances for each node size. Each locker handles pickup and delivery requests for two types of products. The first type weighs 1 kg, and the second type weighs 2 kg. Corresponding delivery stations, paired with lockers, have demand values ranging from $(3, 10]$. For the three customer sizes, the initial truck load is set to 30, 40, and 50 kg, respectively, while the number of no-fly zones is set to 1, 2, and 3, with their locations randomly generated.

\subsubsection{Drone and Truck Parameters}

Each drone is assumed to have a curb weight of 6 kg and a load capacity of 10 kg. The energy consumption per unit is set to 3.5 Wh/(kg$\cdot$km), and the total energy provided by a fully charged drone battery is 56 Wh. For each truck and drone, the fixed costs are taken as \$20 and \$2, respectively. The operating costs are assumed to be \$1.25/km for the truck and \$0.15/km for the drone \cite{Meng2023,Salama2020}.

\subsubsection{Hyperparameters}

Our neural heuristic method is trained using 64 dynamically generated instances during each epoch. The optimal set of hyperparameters is determined as follows: the encoder consists of 6 layers and employs 8 attention heads. The learning rate is set to 0.0001, and the network architecture includes 512 hidden units. Attribute value token embeddings are set to 128 dimensions. The Bi-GRU hidden and cell states are initialized to zero. All training and testing are conducted on a server equipped with an NVIDIA A100 GPU (80 GiB) and AMD Ryzen 9 7945HX CPU at 2.50 GHz. The implementation is developed in Python 3.8 using PyTorch 1.10.

\subsubsection{Baselines}
To comprehensively evaluate our approach, we compare it with several baseline algorithms as follows.

\begin{enumerate}[
    label=(\roman*)
]
\item{Gurobi: An advanced commercial exact solver, used via its Python interface with a maximum time limit of 3600 seconds per instance.}
\item{ALNS \cite{Mulumba2024}: A well-established metaheuristic widely applied to VRPs. Following recent successful applications \cite{Christiaens2020,Wang2025}, we utilize the Slack Induction by String Removals (SISR) destroy-repair operator. This operator is employed both for generating the initial solution and during the iterative search process. Key parameters are configured as follows: the destroy degree is set to 0.2, the maximum number of strings to remove per iteration is 5, and the maximum string size (number of customers per string) is 10. For the associated acceptance criterion, the initial and final temperature control factors, which determine the starting and ending temperatures relative to the objective value of the initial solution, are set to 0.7 and 0.1, respectively. }
\item{PSO \cite{ChandraSugianto2024}: A swarm intelligence algorithm commonly used for solving VRPs. The parameters are configured as follows: the number of particles is 20, the inertia weight ($w$) is 0.8, and both the cognitive factor ($c_1$) and the social factor ($c_2$) are set to 2. The maximum number of iterations is set to 100.}
\item{AM: An attention-based encoder-decoder DRL model. We adapt the implementation settings from \cite{Kool2018} to fit the specific problem structures.}
\end{enumerate}

\subsection{Comparison Study}

For each method, we report the average objective value, optimality gap, and runtime across all test instances. For each problem size, the optimality gap is calculated as:
\begin{equation}
\text{Gap} = \frac{\text{Obj}_{\text{method}} - \text{Obj}_{\text{ref}}}{\text{Obj}_{\text{ref}}} \times 100\%
\end{equation}
where $\text{Obj}_{\text{ref}}$ denotes the best objective value obtained among all feasible methods for the corresponding problem size. When Gurobi obtains the optimal solution within the time limit, its result is used as the reference; otherwise, the best solution obtained by the tested heuristic or neural methods is used as the reference.

\begin{figure}[!t]
\centering
\includegraphics[width=0.7\columnwidth]{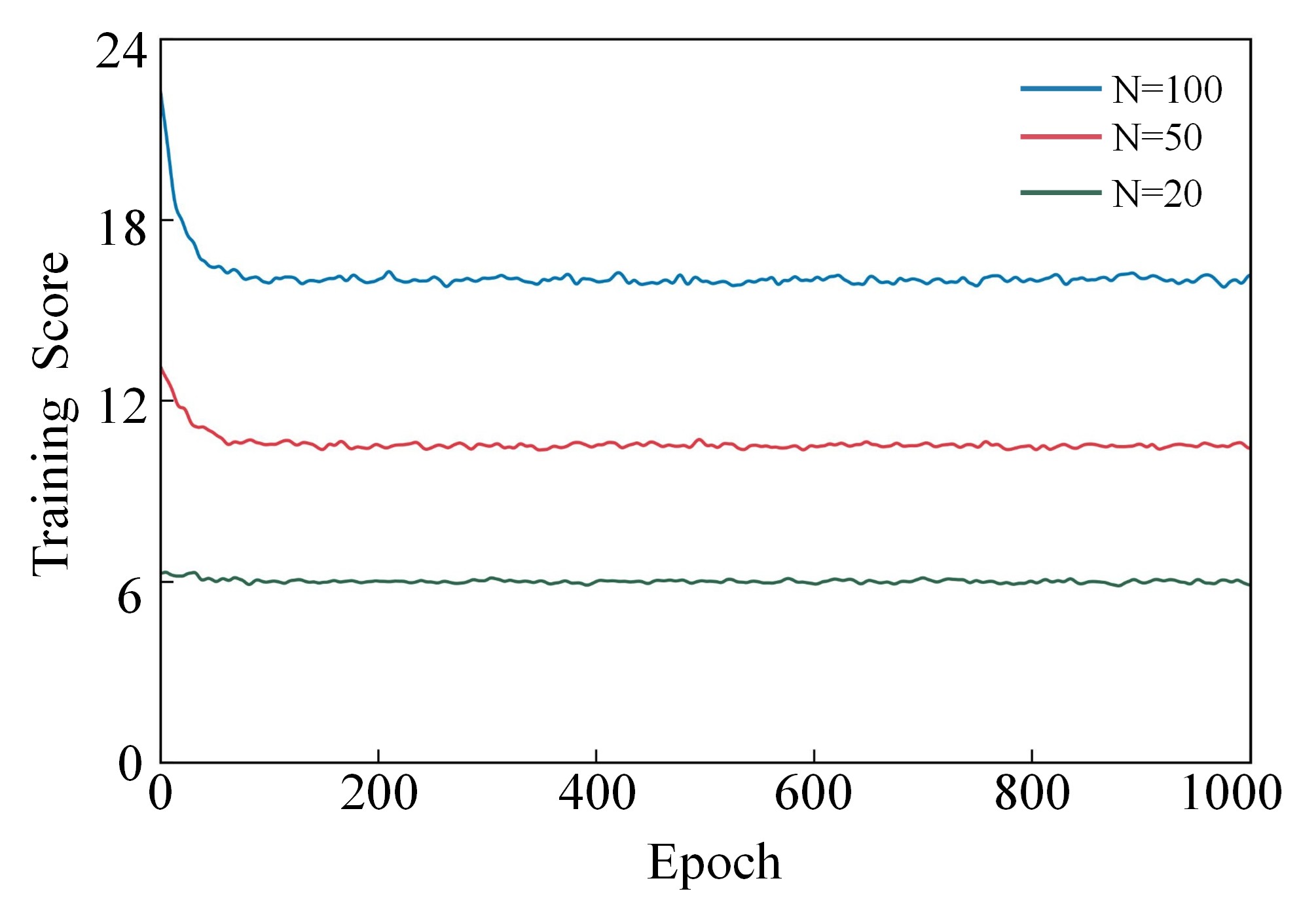}
\caption{Training curves}
\label{fig5}
\end{figure}

Before comparing these methods in detail, we first present the training curves of the DRL method employed in the first stage. Fig. \ref{fig5} displays the training curves for instances with varying node sizes. It is observed that the training process exhibits desirable convergence and stability. For all three node sizes, the training score (i.e., total route length) decreases rapidly at the beginning, after which the training gradually stabilizes.

For CVRP performance evaluation, the objective is to specifically assess the performance of the first stage of our algorithm, which solves the core vehicle routing problem. For LTDRP-PDNF performance evaluation, since our proposed method operates via a sequential two-stage process, we structure the comparison for LTDRP-PDNF consistently across all algorithms to ensure a fair assessment within this specific solution architecture. All baseline methods are therefore also implemented using this mandated two-stage approach. They first determine the truck paths and subsequently, drone paths are constructed according to the established truck routes and coordination constraints. Tables \ref{table3} and \ref{table4} present the comparative results for the CVRP and LTDRP-PDNF, respectively.

\subsubsection{CVRP}
Regarding solution quality, Table III shows that Gurobi obtains the optimal solution for CVRP20, with an average objective value of 6.10. The proposed method achieves an average objective value of 6.14, corresponding to a small gap of 0.66\%, and is closer to the optimum than ALNS, PSO, and AM. For CVRP50 and CVRP100, where Gurobi results are not available, the proposed method obtains the best objective values among all compared methods, with average objective values of 10.46 and 15.80, respectively. Among the baseline methods, AM provides the most competitive solution quality, with gaps of 2.11\%, 1.88\%, and 2.32\% for CVRP20, CVRP50, and CVRP100, respectively. PSO also performs better than ALNS across all three problem sizes, but its solution quality remains inferior to that of AM and the proposed method. ALNS yields the largest gaps among the tested heuristic and neural methods, indicating that its performance is relatively less competitive on the tested CVRP instances.

In terms of computation efficiency, the runtime for all methods generally increases with the number of nodes, as expected. The proposed method achieves the shortest runtime across all problem sizes, requiring only 0.02 s, 0.05 s, and 0.09 s for CVRP20, CVRP50, and CVRP100, respectively. AM also runs efficiently, but its runtime is consistently higher than that of the proposed method. By contrast, PSO and ALNS require substantially longer computation times due to their population-based search and iterative neighborhood search procedures. Therefore, the proposed method achieves the best overall performance in both solution quality and computation efficiency on the CVRP.

\begin{table*}[!!t]
\caption{Results On The CVRP}
\label{table3}
\centering
\begin{tabular*}{\linewidth}{@{\extracolsep{\fill}} l *{9}{c} @{}}
\hline
 & \multicolumn{3}{c}{CVRP20} & \multicolumn{3}{c}{CVRP50} & \multicolumn{3}{c}{CVRP100} \\
\cline{2-4} \cline{5-7} \cline{8-10}  
 & Obj. & Gap & Time(s) & Obj. & Gap & Time(s) & Obj. & Gap & Time(s) \\
\hline
Gurobi & \textbf{6.10} & \textbf{0.00\%} & -- & -- & -- & -- & -- & -- & -- \\
ALNS & 6.26 & 2.62\% & 0.63 & 11.23 & 7.36\% & 1.48 & 17.44 & 10.38\% & 3.37 \\
PSO & 6.25 & 2.38\% & 0.17 & 11.16 & 6.70\% & 0.94 & 17.21 & 8.90\% & 1.77 \\
AM & 6.23 & 2.11\% & 0.05 & 10.66 & 1.88\% & 0.12 & 16.17 & 2.32\% & 0.19 \\
Ours & 6.14 & 0.66\% & 0.02 & \textbf{10.46} & \textbf{0.00\%} & 0.05 & \textbf{15.80} & \textbf{0.00\%} & 0.09 \\
\hline
\end{tabular*}
\end{table*}

\begin{table*}[!!t]
\caption{Results on LTDRP-PDNF}
\label{table4}
\centering
\begin{tabular*}{\linewidth}{@{\extracolsep{\fill}} l *{9}{c} @{}}
\hline
 & \multicolumn{3}{c}{N=20} & \multicolumn{3}{c}{N=50} & \multicolumn{3}{c}{N=100} \\
 \cline{2-4} \cline{5-7} \cline{8-10} 
 & Obj. & Gap & Time(s) & Obj. & Gap & Time(s) & Obj. & Gap & Time(s) \\
\hline
Gurobi & \textbf{77.84} & \textbf{0.00\%} & 160.23 & -- & -- & 3600* & -- & -- & 3600* \\
ALNS & 83.81 & 7.67\% & 7.59 & 158.24 & 7.93\% & 34.59 & 279.73 & 14.80\% & 126.52 \\
PSO & 81.74 & 5.01\% & 9.77 & 157.68 & 7.55\% & 55.98 & 267.28 & 9.69\% & 192.85 \\
AM & 81.52 & 4.73\% & 1.82 & 149.89 & 2.24\% & 6.83 & 251.81 & 3.34\% & 20.10 \\
Ours & 79.96 & 2.78\% & 1.09 & \textbf{146.61} & \textbf{0.00\%} & 3.95 & \textbf{243.67} & \textbf{0.00\%} & 14.79 \\
\hline
\multicolumn{10}{l}{*Denote no feasible solution was found within the specified time limit.} \\
\end{tabular*}
\end{table*}

\subsubsection{LTDRP-PDNF}
Regarding solution quality, Table IV shows that Gurobi obtains the best solution for the small-scale N=20 instances, with an average objective value of 77.84. However, it requires 160.23 s on average and fails to find feasible solutions within the 3600-second time limit for N=50 and N=100. In comparison, the proposed method obtains the second-best solution for N=20, with an average objective value of 79.96 and a gap of 2.78\%, while requiring only 1.09 s. This indicates that the proposed method can achieve near-optimal solution quality for small-scale instances with substantially shorter computation time. For N=50 and N=100, where Gurobi fails to find feasible solutions within the specified time limit, the proposed method obtains the best objective values among all compared methods, with average objective values of 146.61 and 243.67, respectively. AM is the closest baseline, with gaps of 2.24\% and 3.34\% for N=50 and N=100, respectively. PSO performs better than ALNS on the LTDRP-PDNF instances, with gaps of 7.55\% and 9.69\% for N=50 and N=100, whereas ALNS yields larger gaps of 7.93\% and 14.80\%.
In terms of computation efficiency, Gurobi consumes the longest solving time and fails to obtain feasible solutions for N=50 and N=100 within the specified time limit. Among the heuristic and neural methods, the proposed method achieves the shortest runtime across all problem sizes. AM is the second-fastest method, while ALNS and PSO require substantially longer computation times. Overall, the proposed method achieves a favorable balance between solution quality and computation efficiency on LTDRP-PDNF.

\subsubsection{Visualization}

\begin{figure}[!t]
\centering
\subfloat[$N=20$]{\includegraphics[width=0.9\columnwidth]{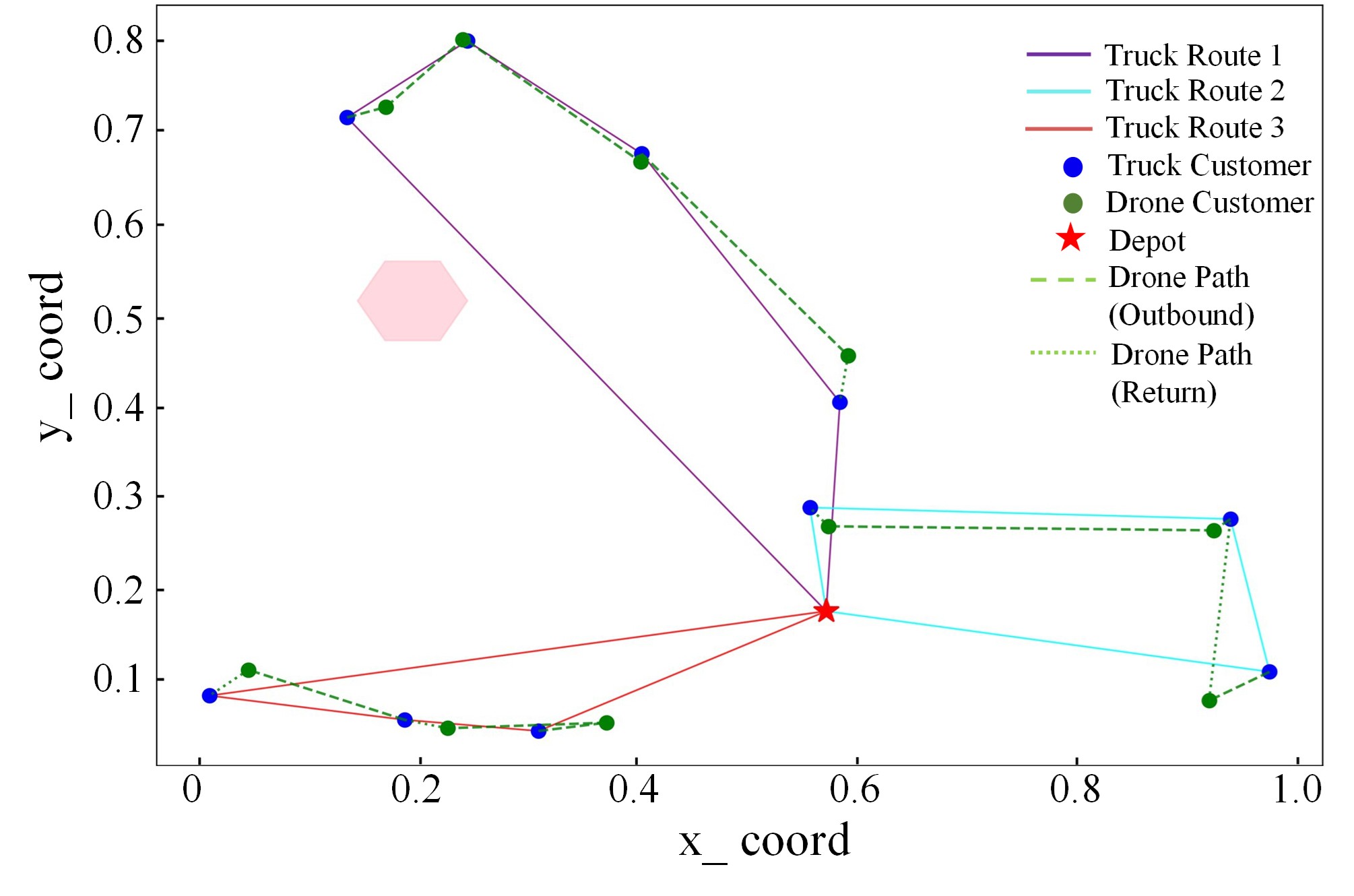}\label{fig6a}}
\\
\subfloat[$N=50$]{\includegraphics[width=0.9\columnwidth]{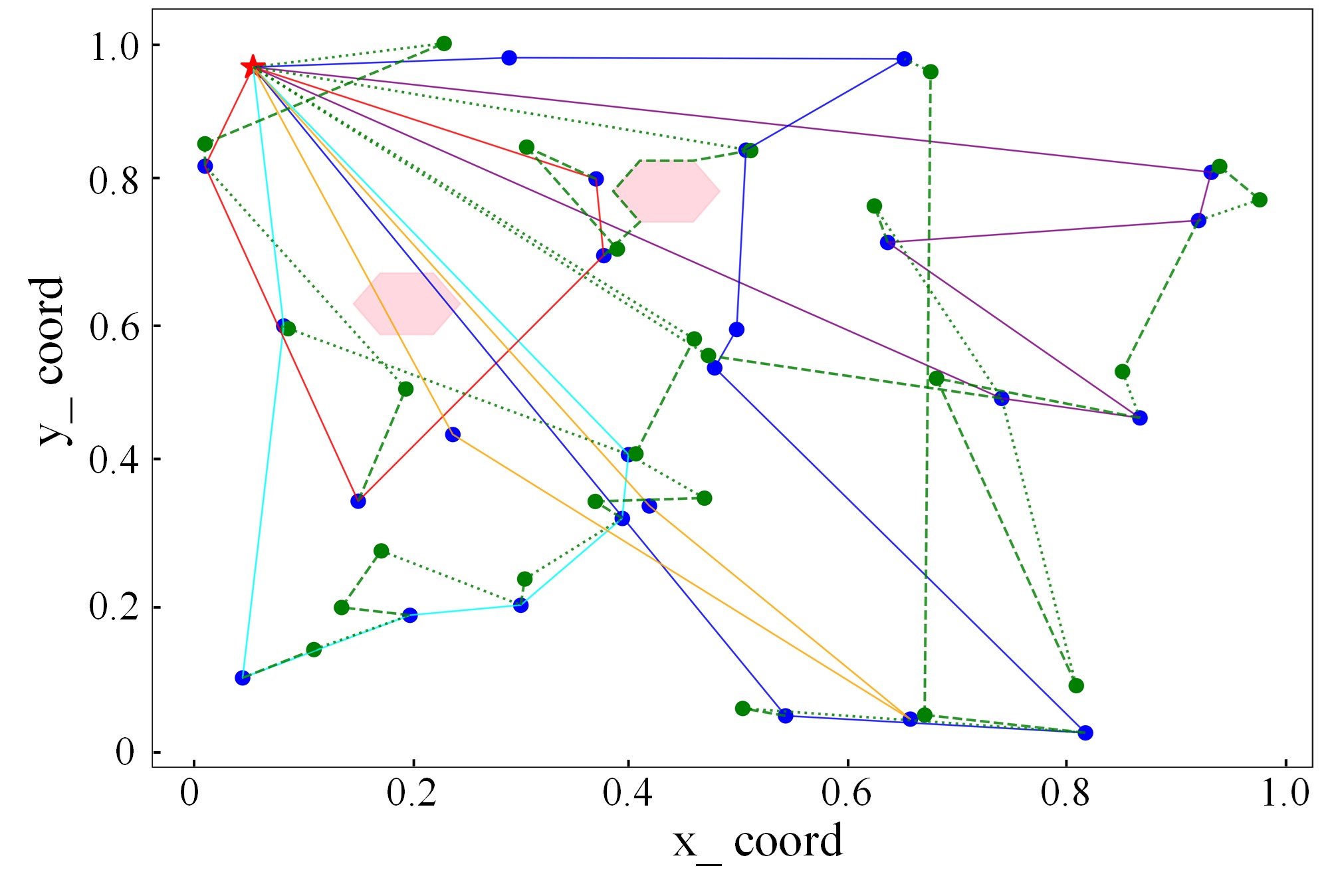}\label{fig6b}}
\caption{Visualization of representative LTDRP-PDNF solutions}
\label{fig6}
\end{figure}

Fig. \ref{fig6} presents two representative solutions generated by the proposed method for LTDRP-PDNF instances with $N=20$ and $N=50$. The red star denotes the depot, blue nodes represent customers served directly by trucks, and green nodes represent customers served by drones. Solid lines indicate truck routes, while dashed lines indicate outbound and return drone flights. The shaded polygons represent no-fly zones. As shown in Fig. \ref{fig6}(a), for the small-scale instance, the truck routes form the main service skeleton, and drones are dispatched from truck-related service points to cover nearby drone customers. The generated drone paths successfully avoid the no-fly-zone while maintaining feasible connections with the truck routes. Fig. \ref{fig6}(b) further illustrates the solution structure for a larger instance, where more truck and drone service nodes are involved and the route network becomes more complex. Even under this increased scale, the proposed method can construct coordinated truck–drone routes while considering locker-based service, drone flight feasibility, and no-fly-zone avoidance.

\subsection{Generalization Study}
To evaluate the generalization ability of our neural heuristic method, we further evaluate our method on different cross-size instances. Specifically, we generated three sets of larger-scale LTDRP-PDNF instances, with the number of nodes set to 30, 60, and 110, respectively. Given the low computational efficiency of Gurobi, we only compare our method with ALNS, PSO, and AM.

\begin{figure}[!t]
\centering
\includegraphics[width=0.7\columnwidth]{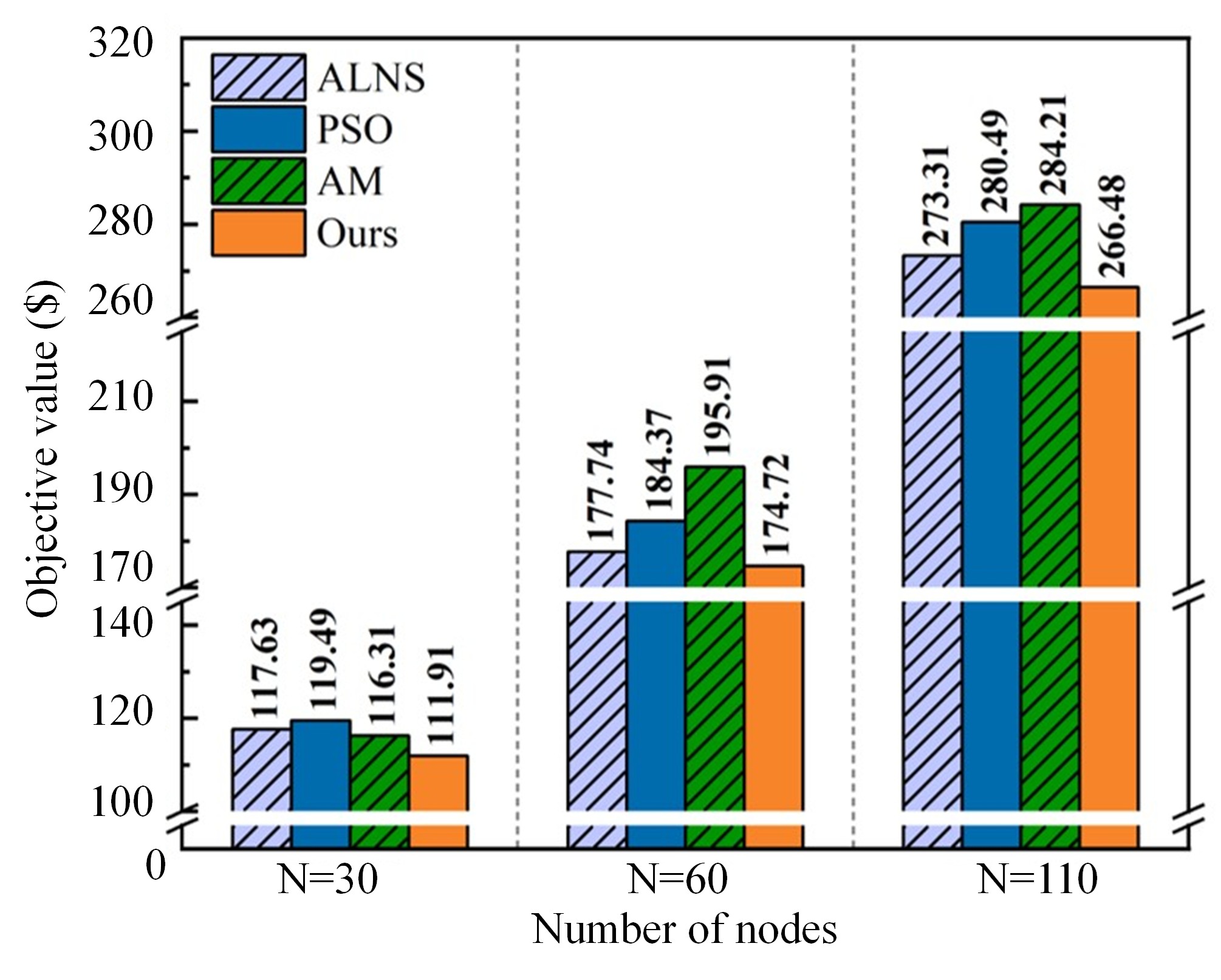}
\caption{Generalization to different sizes}
\label{fig7}
\end{figure}

The average total cost results are presented in Fig. \ref{fig7}. Empirical results show that our method exhibits superior generalization capability, achieving the lowest total cost when solving instances with larger node sizes. While ALNS and PSO perform reasonably well, their solution quality remains significantly inferior to ours. Despite the competitive performance of AM, our neural heuristic is able to further enhance solution quality over it, which verifies the effectiveness of our method.
\subsection{Sensitivity Analysis}
In this section, we investigate the sensitivity of the total solution cost to two critical drone parameters: maximum load capacity and battery capacity. Our analysis focuses specifically on these two factors, examining their effects across different locker layout configurations. These configurations are defined by the ratio of delivery stations to lockers (denoted D/L), evaluated for D/L values of {1, 2, 3}. To elucidate these impacts, we conduct evaluations on the LTDRP-PDNF dataset with 50 nodes, which includes 100 instances. For clarity and comparability, each experiment modifies only one parameter at a time while keeping all other parameters fixed. This sensitivity analysis was conducted separately for each D/L ratio. Throughout these experiments, the fixed cost per drone was consistently set to half the fixed cost per truck.

\begin{figure}[!t]
\centering
\includegraphics[width=0.6\columnwidth]{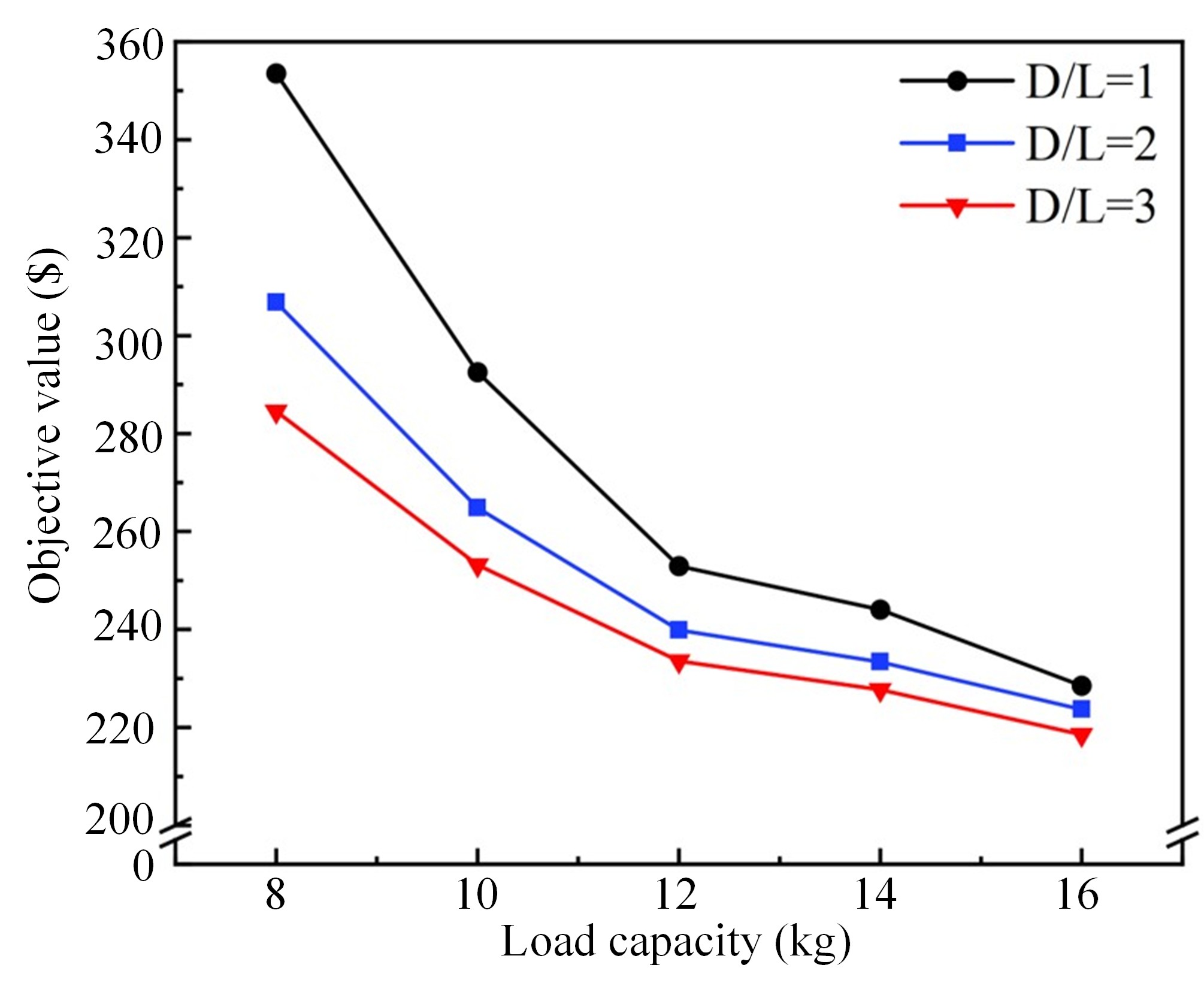}
\caption{Impact of drone maximum load capacity on total cost}
\label{fig8}
\end{figure}

\begin{figure}[!t]
\centering
\includegraphics[width=0.6\columnwidth]{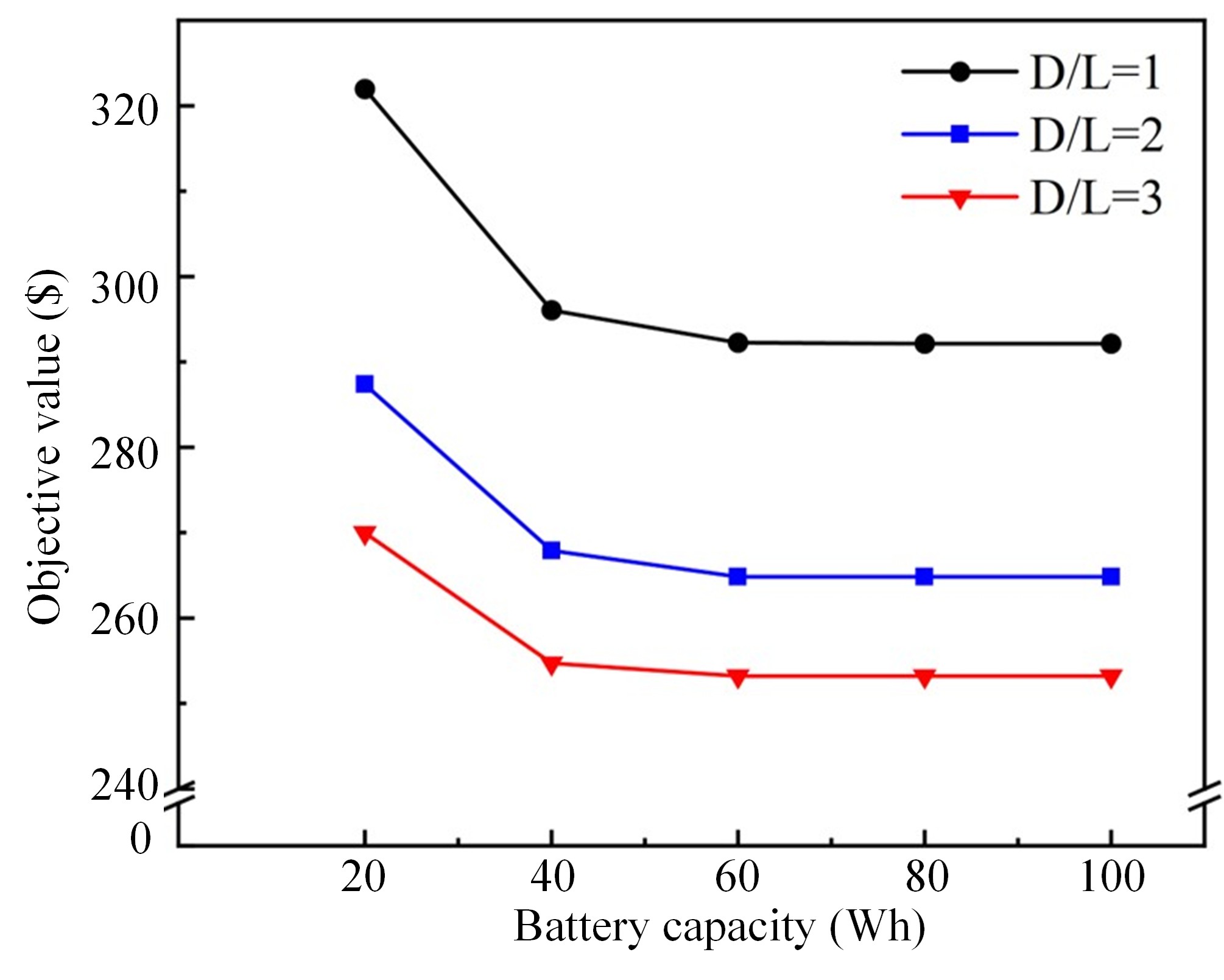}
\caption{Impact of drone battery capacity on total cost}
\label{fig9}
\end{figure}

Fig. \ref{fig8} illustrates the effect of increasing drone load capacity on the total delivery cost. Generally, as load capacity rises, the total cost decreases across all layout configurations (D/L ratios), and the cost differences between these layouts also tend to narrow. This suggests that moderate improvements in drone payload enable each drone to serve more lockers per trip, thereby reducing the number of drones required and lowering the overall cost. However, there is a threshold in load capacity beyond which the rate of decrease becomes slower. This is because, in order to minimize delivery costs, each truck typically dispatches drones only to relatively nearby lockers along its route. Consequently, even with ample capacity, a single drone is unlikely to serve a vast number of lockers. Once the drone load capacity surpasses a certain threshold, further increases provide limited additional benefit to the total cost.

In contrast, Fig. \ref{fig9} reveals the impact of increasing drone battery capacity. Here, the total cost drops sharply at first, followed by a more gradual decline, and eventually stabilizes. 
This is primarily because greater battery range allows drones to access more distant lockers from the truck route. While serving farther lockers increases the travel cost per drone flight, it can enable a significant reduction in the total number of drone deployments, leading to substantial overall cost savings. However, since both trucks and lockers are capable of replacing drone batteries, drones do not require excessively large battery capacities to complete their delivery tasks. In this case, load capacity becomes the primary limiting factor, preventing drones from serving more lockers even if higher battery capacity is available.
The sensitivity analysis clearly demonstrates that both the load capacity and battery capacity of drones have substantial impacts on the total operating cost. Consequently, it is crucial to carefully consider these factors in practical operations. It is clear from the results that appropriately tuning these parameters under different delivery station-to-locker ratios can yield significant reductions in delivery costs.

\section{Conclusion}

This paper introduces LTDRP-PDNF, motivated by the practical need for integrated coordination among trucks, drones, and parcel lockers in last-mile delivery operations. The proposed framework jointly considers multi-product demands, simultaneous pickup and delivery, payload-dependent drone energy consumption, no-fly-zone detours, and locker-assisted battery replacement, thereby providing a more realistic representation of real-world truck–drone delivery systems than conventional routing models.

To solve this challenging NP-hard problem, we formulate the route construction process as an MDP and develop a two-stage DRL-based neural heuristic. The proposed framework first learns high-quality truck-routing policies from the CVRP and subsequently transfers the acquired knowledge to construct coordinated truck–drone solutions for LTDRP-PDNF. Computational experiments demonstrate that the proposed approach consistently achieves superior solution quality and computational efficiency compared with both conventional heuristic and learning-based benchmark methods. In addition, the method exhibits strong scalability and generalization capability across different problem sizes.

Managerial insights are further obtained through sensitivity analyses. The results indicate that drone payload capacity and battery capacity significantly influence the overall delivery cost, highlighting the importance of balancing vehicle configuration decisions with operational planning. These findings provide useful guidance for logistics operators seeking to improve the efficiency and cost-effectiveness of locker-based truck–drone delivery systems.

Several promising research directions deserve further investigation. First, dynamic no-fly zones, such as time-dependent flight restrictions and temporary airspace regulations, could be incorporated to better reflect realistic operating environments. Second, future studies may consider time windows, stochastic customer requests, and dynamic order arrivals to enhance the practical applicability of the model. Finally, more advanced reinforcement learning techniques and transfer-learning strategies could be explored to further improve solution quality, training efficiency, and cross-scale generalization performance.

\newpage

\bibliographystyle{IEEEtran}
\bibliography{reference}

% \clearpage
% \raggedbottom

\begin{IEEEbiography}
[{\includegraphics[width=1in,height=1.25in,clip,keepaspectratio]{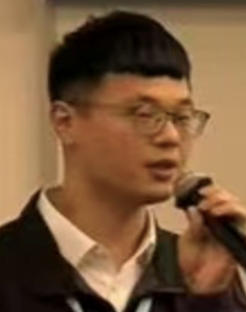}}]{Xuanyu Liu}
is currently pursuing the Ph.D. degree with Chang’an University. He received the B.S. degree from Zhengzhou University of Aeronautics in 2023. His current research interests include delivery system design and truck-drone collaborative routing optimization.
\end{IEEEbiography}

\begin{IEEEbiography}
[{\includegraphics[width=1in,height=1.25in,clip,keepaspectratio]{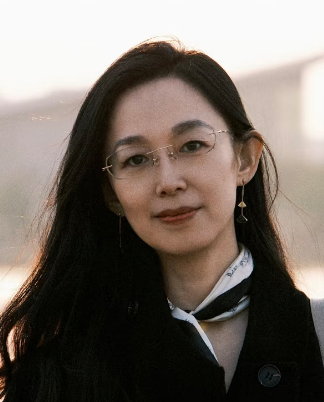}}]{Hui Hu}
received the Ph.D. degree in systems engineering from Beijing Jiaotong University, Beijing, China, in 2008. She is currently a professor and the Director of the Institute of Transportation Systems Organization and Control, Chang’an University, China. She was a visiting scholar at the University of Wisconsin–Madison, WI, USA, and Eindhoven University of Technology, The Netherlands.
Her research interests include transportation planning, logistics optimization, intelligent logistics systems, and truck–drone collaborative delivery.
She has authored or coauthored more than 40 publications and also contributed to the development of a national standard in the field of logistics and transportation. Her professional services include serving as a Board Member of the China Logistics Society, a Committee Member of the World Transport Convention (WTC), and a Think Tank Expert for the Shaanxi Provincial Development and Reform Commission.
\end{IEEEbiography}

\vspace{-10mm}

\begin{IEEEbiography}
[{\includegraphics[width=1in,height=1.25in,clip,keepaspectratio]{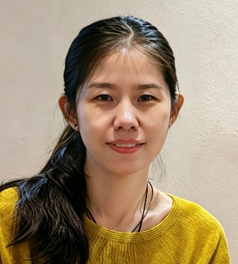}}]{Jiao Zhao}
received her B.S. degree and M.S. degree from Chang'an University, and her PhD degree in system engineering from Northeastern University in 2013. She is currently an associate professor in the College of Transportation Engineering at Chang'an University. She has authored or coauthored more than 20 publications. Her current research interests include port scheduling optimization, transportation planning and logistics optimization.
\end{IEEEbiography}

\vspace{-10mm}

\begin{IEEEbiography}
[{\includegraphics[width=1in,height=1.25in,clip,keepaspectratio]{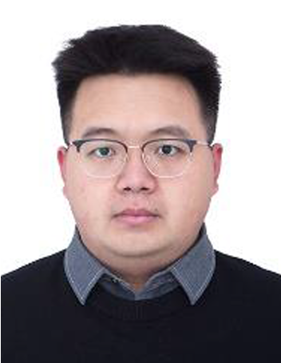}}]{Ziliang Wang}
is a Lecturer from School of Transportation Engineering in Chang’an University. He received the B.S. and Ph.D. degrees from Northwestern Polytechnical University in 2019 and 2024, respectively. He has published articles in leading international journals, including IEEE Transactions on Intelligent Transportation Systems, International Journal of Production Research, etc. His current research interests include transportation planning and optimization, and operations research.
\end{IEEEbiography}

\vspace{-10mm}

\begin{IEEEbiography}[{\includegraphics[width=1in,height=1.25in,clip,keepaspectratio]{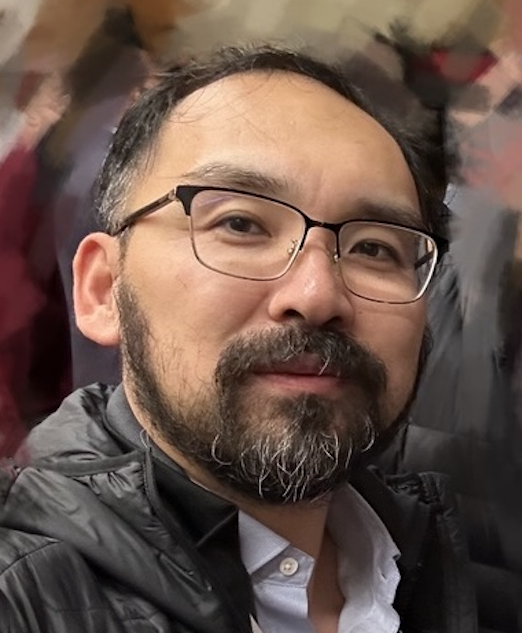}}] 
{Zhengbing He} (M'17-SM'20) received the Bachelor of Arts degree in English language and literature from Dalian University of Foreign Languages, China, in 2006, and the Ph.D. degree in systems engineering from Tianjin University, China, in 2011. 
He was a Postdoctoral Researcher and an Assistant Professor with Beijing Jiaotong University, China. From 2018 to 2022, he was a Professor at Beijing University of Technology, China. 
From 2023 to 2025, he was a research scientist with Massachusetts Institute of Technology, USA. 
Presently, he is a Professor with University of Nottingham Ningbo China. 

\vspace{-2mm}

His research lies at the intersection of urban mobility, systems engineering, and artificial intelligence, spanning from traditional topics such as traffic flow operations and control, sustainability, and resilience to emerging areas including data-driven modeling, autonomous driving, and large language models. In particular, he is a pioneer and long-term contributor in AI-based transportation modeling and AV-empowered traffic congestion mitigation, and an early innovator in generative AI applications in transportation.

\vspace{-2mm}

He has published more than 180 papers, including over 50 published exclusively in the prestigious journal series of Transportation Research and IEEE TRANSACTIONS (20+ TRC and 10+ TITS) and a Correspondence in Nature, with total citations exceeding 8,000 and H-index of over 40.
He was listed as the World’s Top 2\% Scientists, ranking 67th out of over 30,000 researchers in the field of Logistics and Transportation. He is the Editor-in-Chief of the Journal of Transportation Engineering and Information (Chinese). Meanwhile, he serves as a Senior Editor for IEEE TRANSACTIONS ON INTELLIGENT TRANSPORTATION SYSTEMS, an Associate Editor for IEEE TRANSACTIONS ON INTELLIGENT VEHICLES, a Deputy Editor-in-Chief of IET Intelligent Transport Systems, and an Editorial Advisory Board Member for Transportation Research Part C. His webpage is https://www.GoTrafficGo.com.
\end{IEEEbiography}

\end{document}